\documentclass[twoside,twocolumn,10pt]{article}

\usepackage{fancyhdr}
\usepackage{geometry}
\usepackage{abstract}
\usepackage{graphicx}
\usepackage{titlesec}
\usepackage{ragged2e}
\usepackage{amssymb}
\usepackage{subcaption}
\usepackage{parskip}
\usepackage{amsmath}
\usepackage{newtxtext}
\usepackage{booktabs}
\usepackage{array}
\usepackage{balance}
\usepackage[backend=bibtex,style=ieee]{biblatex}
\usepackage[font=footnotesize,labelfont=bf,justification=raggedright,format=plain]{caption}
\usepackage{float}

\defbibheading{bibliography}[\refname]{%
  \section{\MakeUppercase{REFERENCES}}}

\newcolumntype{L}[1]{>{\raggedright\arraybackslash}p{#1}}
\newcolumntype{C}[1]{>{\centering\arraybackslash}p{#1}}
\newcolumntype{R}[1]{>{\raggedleft\arraybackslash}p{#1}}

\fancypagestyle{firstpage}{
  \fancyhf{}
  \fancyhead[LO]{\small August 2026}
 \fancyhead[RO]{\small The University of Osaka, Computer Science and Computer Vision}

}

\title{\fontsize{16}{20}\selectfont\textbf{PlantRig -- From Bones to Branches: Adaptation of Autoregressive Rigging Models for Plant Skeletal Reconstruction}}

\author{
    \fontsize{12}{14}\selectfont
    Nathan Hu\textsuperscript{1*}, Yang Yang\textsuperscript{2}, Fumio Okura\textsuperscript{3}\\[1ex]
    \fontsize{12}{14}\selectfont
    \textsuperscript{1}Mathematics, UCLA, Los Angeles, CA, United States of America \\
    \textsuperscript{2}Computer Science, University of Osaka, Osaka, Japan \\
    \textsuperscript{2}Computer Science, University of Osaka, Osaka, Japan
}

\date{}

\titleformat{\section}{\normalsize\bfseries\uppercase}{\thesection.}{1em}{}
\titleformat{\subsection}{\normalsize\bfseries\flushleft}{\thesubsection.}{1em}{}
\titleformat{\subsubsection}{\normalsize\bfseries\itshape\flushleft}{\thesubsubsection.}{1em}{}

\begin{document}

\twocolumn[
\begin{@twocolumnfalse}

\vspace{-0.7cm}
\noindent
\raggedright
{\fontsize{11}{13}\selectfont Original Article and Manuscript}
\hfill

\vspace{-0.5cm}

\maketitle
\thispagestyle{firstpage}

\section*{\hspace{-1.5mm}Abstract}
\fontsize{10}{12}\selectfont
\justifying
\noindent
Autoregressive rigging models such as UniRig and SkinTokens perform well on articulated characters, but their ability to generalize to plant structures remains largely unexplored, since plant topologies exhibit highly variable, non-canonical branching patterns that challenge learned skeletal priors. We evaluate these models for plant skeletal reconstruction using synthetic L-system-generated trees and real scanned data spanning monopodial, sympodial, whorled, and vine-like archetypes. Preliminary testing showed UniRig collapsing complex branching into near-linear chains, while SkinTokens preserved topology better but over-segmented branches and produced an unstable output space, so we focused on UniRig for its greater stability. Diagnosis traced the collapse to sampling-level suppression of branch tokens, and further analysis showed the frozen mesh encoder had limited sensitivity to structural variation, pointing to a geometric bottleneck in the tokenization pipeline rather than a purely learned bias. Building on these findings, we applied multi-round fine-tuning over multiple procedurally generated synthetic datasets. Across rounds, the model progressively recovered accurate branching topology and generalized beyond branch-only structures to plants with foliage, a harder case given the zero-thickness, mesh-normal-dependent geometry of leaves. The resulting model generalized well across diverse plant forms without leaf-specific architectural changes, indicating that targeted fine-tuning can substantially close the domain gap between character-rigging priors and plant skeletal structure. As such, our work points toward a viable path for automated plant rigging across both branch topology and foliage type, even those not considered in our findings.

\vspace{0.3cm}
\noindent\textbf{Keywords:}
Autoregression, Rigging, Plants, Skeletal Estimation, Optimization, Supervised Learning, Deep Learning, Computer Vision, Ecology, Python

\vspace{0.8cm}
\end{@twocolumnfalse}
]

\section{Introduction} 
\fontsize{11}{13}\selectfont
Plants play a fundamental role in terrestrial ecosystems, regulating global carbon cycling, supporting biodiversity, influencing climate, and providing essential resources for agriculture and forestry. Understanding plant structure is therefore critical for applications such as high-throughput phenotyping, precision agriculture, robotic pruning and harvesting, biomass estimation, growth modeling, and the development of agricultural digital twins \cite{ref8}. Central to many of these applications is the recovery of a plant's skeletal structure, which provides a compact representation of its branching topology while preserving the hierarchical relationships that govern its growth.

Traditional plant skeletal reconstruction methods have largely relied on geometric optimization and handcrafted structural constraints to extract branches from three-dimensional point clouds \cite{ref1,ref2}. More recently, learning-based approaches have demonstrated improved robustness by directly predicting skeletal graphs from observed geometry \cite{ref3,ref7}. However, these methods generally do not take advantage of recent advances in universal skeleton generation.

In computer graphics, skeleton rigging has become a fundamental technique for representing the hierarchical structure of three-dimensional objects. Beyond enabling animation, rigged skeletal representations provide an efficient framework for describing topology, structural relationships, and deformation. Recent autoregressive transformer-based rigging models such as UniRig and SkinTokens have demonstrated remarkable performance by learning to generate skeletal structures directly from three-dimensional geometry across many object categories \cite{ref4,ref5}. Despite this progress, their application to plants remains largely unexplored yet natural, due to the recursive branching patterns, variable connectivity, and topological diversity that distinguish botanical structures from conventional articulated objects.

This study investigates the capacity of these models, specifically UniRig, to reconstruct plant skeletal structures from 3D mesh data. Furthermore, it examines the causes of branching deficiencies observed during baseline inference and evaluates which modifications most effectively improve reconstruction quality for plant-specific topologies. The project also provides scripts to deal with the aforementioned problems, including an archetypal plant generation script and an autoencoding mesh script. It also proposes a novel segmentation point of view that tokenizes foliage-based models.

\section{Literature Review}
\fontsize{11}{13}\selectfont
\subsection{Plant Skeletal Extraction}
Recovering skeletal representations from three-dimensional plant data has traditionally been formulated as a geometric optimization problem. One of the most influential approaches was proposed by Chaudhury and Godin \cite{ref1}, who introduced a stochastic optimization framework for extracting skeletons from plant point clouds. Their method begins by representing the point cloud as a probabilistic model and iteratively refines skeletal points through expectation-maximization while enforcing smoothness and connectivity. Unlike purely geometric thinning algorithms, the optimization explicitly seeks the medial structure of the plant while remaining robust to moderate levels of scanning noise and missing observations. The resulting skeletons preserve branch continuity more effectively than earlier heuristic methods. However, the framework relies on carefully designed objective functions and initialization procedures, making performance sensitive to parameter selection and increasing computational cost for highly complex and diverse branching structures.

Wu et al. \cite{ref2} addressed similar challenges in maize reconstruction by exploiting biological characteristics specific to crop plants. Their algorithm combines neighborhood analysis, graph construction, and branch refinement to generate central lines that closely follow the underlying stem architecture. By incorporating species-specific assumptions regarding branching patterns, the method achieves high accuracy for maize phenotyping applications, particularly in recovering stem geometry from terrestrial laser scans. Nevertheless, the reliance on handcrafted rules limits its applicability to other plant species exhibiting significantly different topologies, such as shrubs, trees, or ornamental plants with highly recursive branching.

These optimization-based methods established that explicitly modeling plant topology substantially improves reconstruction quality compared with purely geometric surface analysis. However, both approaches remain dependent on manually engineered objectives and priors, making generalization across diverse plant morphologies difficult. Furthermore, as plant complexity increases, optimization becomes progressively more expensive, motivating the exploration of learned representations.

A significant shift toward learning-based reconstruction is demonstrated by Smart-Tree \cite{ref3}, which replaces explicit optimization with a neural approximation of the plant's medial axis. Instead of iteratively searching for skeletal centerlines, Smart-Tree trains a neural network to infer them directly from point clouds. This substantially improves robustness to irregular point densities, occlusions, and incomplete observations commonly encountered in outdoor scanning environments. The learned representation also allows inference to be considerably faster than iterative optimization methods once training is complete. Despite these advantages, Smart-Tree remains focused primarily on predicting skeletal geometry rather than learning a generative representation of skeletal topology. Consequently, its ability to generalize across fundamentally different branching organizations remains dependent on the diversity of the training dataset.

\subsection{Procedural Plant Generation}
A fundamental challenge in developing learning-based reconstruction algorithms is obtaining sufficiently large datasets containing accurate ground-truth skeletons. Procedural modeling provides an effective solution to this problem. The practice introduced by Prusinkiewicz and Lindenmayer in \textit{The Algorithmic Beauty of Plants} \cite{ref6} remains the foundation of procedural plant generation. Lindenmayer systems (L-systems) describe plant development through recursive, mathematical branching rules, allowing complex labyrinths to emerge from compact symbolic rules.

One of the most important characteristics of L-systems is their explicit encoding of topology. Every branch, junction, and parent-child relationship is generated directly through grammar production, providing exact skeletal information alongside the resulting geometry. Unlike skeletons extracted from scanned data, these procedural skeletons are free from measurement noise and ambiguity. While this may lead to a larger real-to-sim gap, one can simply add a layer of autoencoded surface noise that mimics collection of real-world data. This yields a highly realistic synthetic data source, capable of simulating vegetation. Furthermore, production rules can be systematically modified to generate a wide range of botanical architectures, including monopodial trees, sympodial growth, palms, shrubs, and herbaceous plants. This control has made L-systems a valuable source of synthetic training data for computer vision tasks where annotated real-world datasets are scarce or impossible to expand.

\subsection{Learning-Based Plant Reconstruction}
Recent work has increasingly incorporated deep learning into plant structural reconstruction. PlantPose \cite{ref7} formulates skeletal estimation as a graph generation problem rather than a purely geometric task. The method introduces tree-constrained graph generation to ensure that predicted skeletons satisfy practical connectivity constraints. Rather than independently predicting skeletal joints, PlantPose simultaneously learns node positions and graph connectivity, reducing the occurrence of disparate branches and topological inconsistencies. This represents an important transition from geometry-centric reconstruction toward topology-aware learning.

Similarly, GaussianPlant \cite{ref9} explicitly models plant frameworks, which improves three-dimensional reconstruction \textit{quality}. Instead of treating Gaussian splats as independent primitives, the method aligns Gaussian representations with the underlying branching architecture, producing reconstructions that better preserve thin stems and complex plant geometry. The work illustrates that structural priors remain valuable even within modern neural rendering frameworks, reinforcing the importance of topology-aware representations throughout the reconstruction pipeline.

Although these methods significantly improve structural prediction, they remain specialized for plant reconstruction tasks and do not investigate transferable skeletal representations that generalize beyond static representations.

\subsection{Autoregressive Skeletal Generation}
Parallel developments within computer graphics have produced powerful autoregressive methods for rigging. UniRig \cite{ref4} introduces a unified transformer architecture capable of sequentially predicting complete skeletal rigs for many groups of objects. Instead of relying on handcrafted rigging heuristics, the model generates skeletal joints autoregressively, learning both its topology and spatial relationships from large collections of rigged meshes. The sequential formulation naturally captures hierarchical dependencies between joints, allowing the network to construct coherent skeletal structures without explicitly programmed rules.

SkinTokens \cite{ref5} extends this autoregressive paradigm by learning a compact latent representation that jointly encodes all of the above alongside skinning weights. Rather than predicting skeletal geometry alone, the model tokenizes rigging information into discrete learned embeddings that are generated autoregressively using transformer architectures. This unified representation simplifies the rigging pipeline while demonstrating that complex hierarchical structures can be represented as sequential prediction problems. The work further illustrates the scalability of transformer-based rigging methods across heterogeneous object categories.

Despite their impressive performance, both UniRig and SkinTokens were developed primarily for articulated objects such as humans, animals, and mechanical models. These domains possess relatively constrained skeletal topologies characterized by fixed limb connectivity and limited branching factors. In contrast, plants exhibit recursive growth, variable branching order, self-similarity across scales, and topological diversity that differs fundamentally from articulated skeletons. Consequently, it remains unclear whether autoregressive rigging architectures trained on conventional rigging datasets can effectively model botanical structures without architectural modifications or domain adaptation.

\section{Data Methodology}
\fontsize{11}{13}\selectfont
\subsection{Data Collection}
Collecting plant-based meshes to use would not be a sinecural task. Moreover, it would be difficult to find properly cultivated plants and bring them into a controlled environment in short notice to collect complete three-dimensional models to use. The process is both prone to noise and expensive, and would lead to less thorough inference due to the prolonged data collection process. Another thing to note is that for thorough testing, we needed a diverse sample of genuses, some of which do not grow at all in our lab's environment, which would severely limit our project's applications had we only used existing species nearby. This doesn't mean that we did not use real-life data, though, as the reader will see. 

First, for the initial evaluation of UniRig, we look to sister projects in the field of plant-based, ecological computer vision. We looked at and reused data from the GaussianPlant project \cite{ref9}, some of which are present in the form of dense point clouds as a result of three-dimensional Gaussian splatting. With existing techniques, specifically ball-pivoting surface reconstruction due to its preservation of thinner spatial details, we were able to convert these into mesh files. UniRig takes these as its input. This gives us branch-only data in which we focus on reconstructing an accurate \textit{skeleton} first, the baseline before any weights can be applied to extend the autoregression to include foliage. Moreover, for later visualization and higher levels of inference, we borrowed GaussianPlant's meshes of full plants, taken from their \textit{two-dimensional} Gaussian splatting pipeline, as well as meshes of a real life soybean plant.

\begin{figure} [h]
    \centering
    \includegraphics[width=0.8\linewidth]{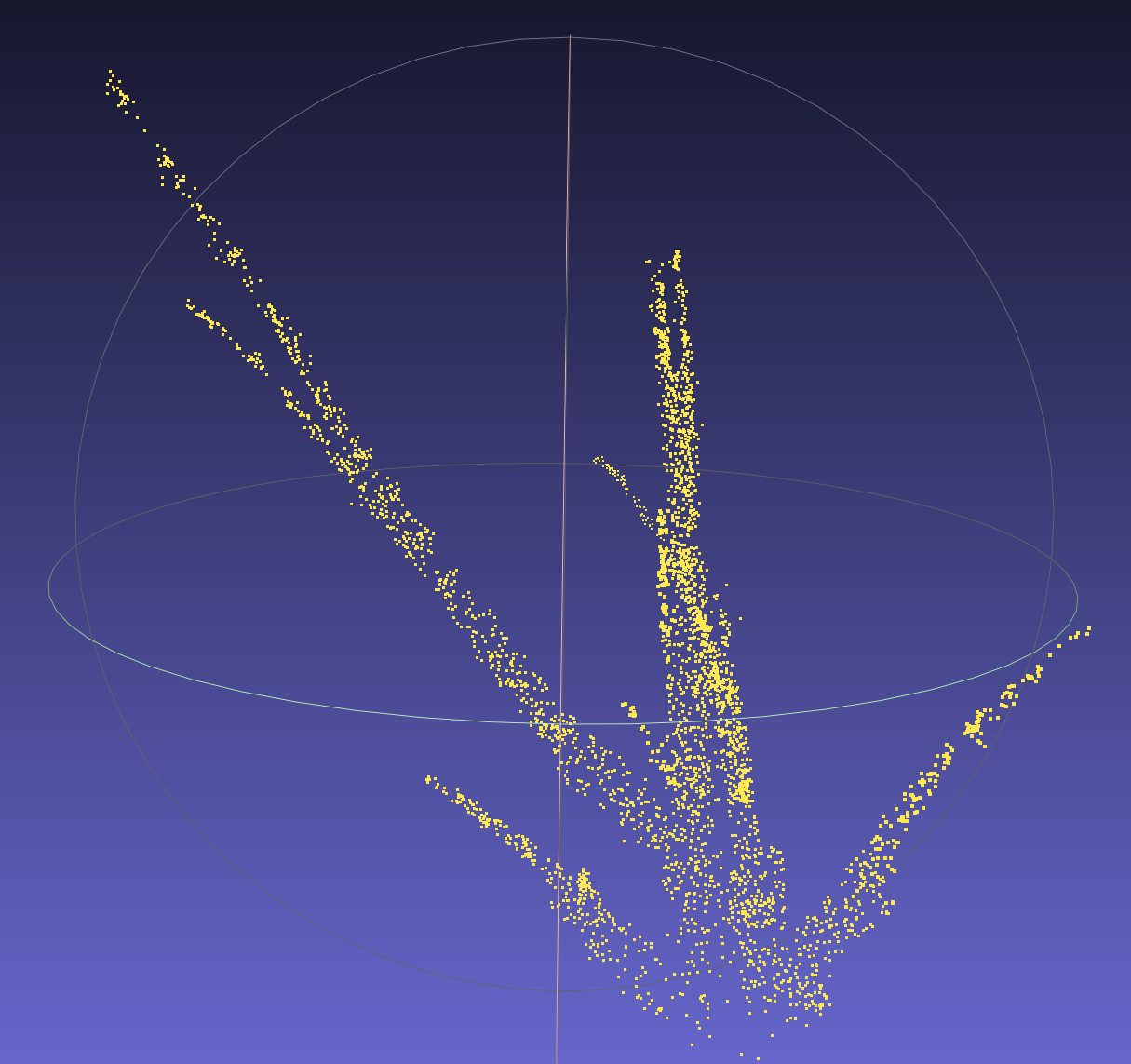}
    \caption{Example of a dense point cloud after GaussianPlant's \cite{ref9} three-dimensional Gaussian splatting experiment}
    \label{fig:placeholder}
\end{figure}
\begin{figure} [h] 
    \centering
    \includegraphics[width=0.9\linewidth]{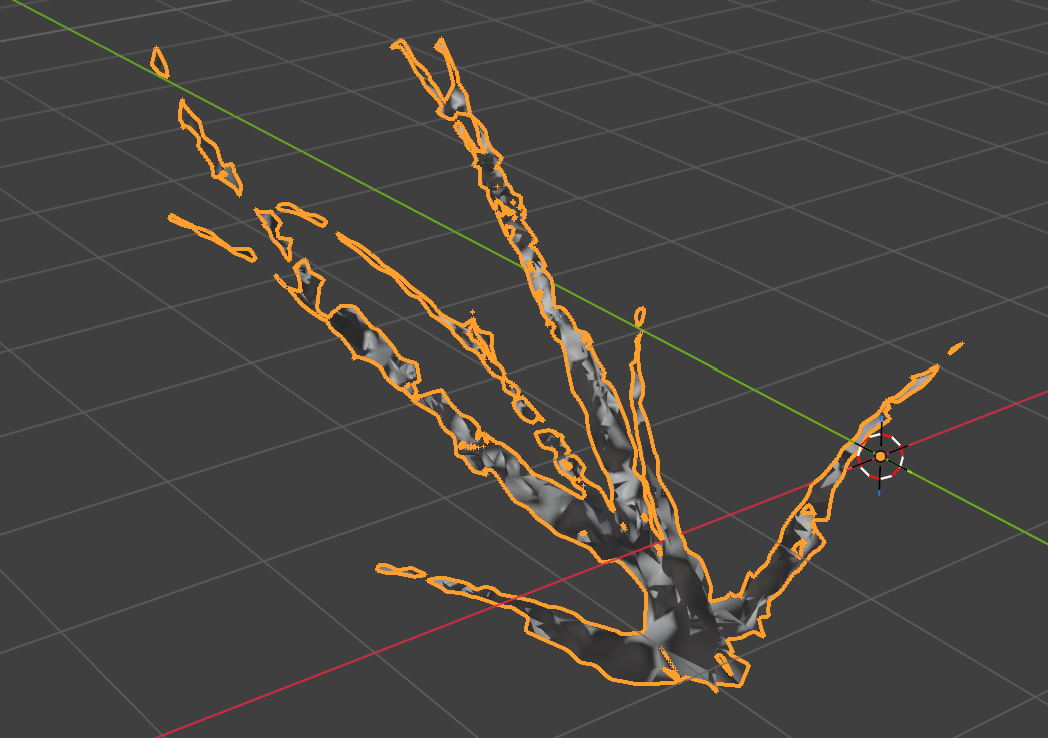}
    \caption{Mesh from real plant data (same plant as in Figure 1), dense point cloud and ball-pivoting surface reconstruction; taken from GaussianPlant \cite{ref9}}
    \label{fig:placeholder}
\end{figure}

Obviously, mesh conversion techniques are not nearly perfect. Thus, to help with our project, we needed a noiseless baseline to help with both initialization and fine-tuning. Such a set would also assist us in evaluating the model's performance and determining if any shortcomings we observed were due to noise. As such, we turned to Lindenmayer systems, which directly and procedurally generate 3D meshes for us to use, via a mathematical system of logic-based branching rules. There are many resources on the internet that accomplish this goal, but none of them have the ability to generalize across many families of plants. Hence, we decided to create our own novel system, one that encompasses a larger input space to expose to UniRig. Due to this input space , these meshes would be perfect for our setting. Moreover, the original usage of such techniques were already for targeting botany, specifically to model \textit{plant} growth and morphology. This directly aligns with our premise, as we exactly strive for realistic yet 3D plant meshes.

Lastly, for our final evaluation---rigging an entire plant including its foliage---we wrote Python scripts that were able to expand upon the branch-only, confined structures from above. We added leaves and minor appendages like petioles into the mesh that weren't included as part of the ground truth skeleton, which we were able to use for training the model further. Within the addition of lesser limbs, we also included functionality to add surface noise to the generated meshes, mimicing an artificial noisy environment that one might observe with real-life scanning techniques such as LiDAR scanning. However, simply relying on this kind of a noise to bridge the real-to-sim gap would not be enough, and so we scripted another Python package that features an noise autoencoder of sorts, detailed later. Lastly, to make the meshes even more lifelike, varying attributes were added to adjust accessory components such as foliage density, individual leaf thickness, as well as unique leaves for certain archetypes (as with palm trees and cycads having fronds, for instance). Once we figured out how to segment and tokenize the plant mesh without just isolating the branches, this synthetic data proves to be a great means of evaluation and adjustment, encompassing a broad array of species that one might not have access to.

\subsection{Data Preprocessing}
The preprocessing stage applies to two distinct categories of our input data, each requiring a different pipeline before they can be used for training and evaluation. 

The first category consists of point cloud data as mentioned above from our sister project, GaussianPlant, which represent plant geometry as a collection of discrete three-dimensional points. While point clouds accurately capture the external shape of a plant, they do not inherently contain surface connectivity or mesh topology, both of which are required by the autoregressive rigging models used in our work. To bridge this gap, the point clouds were converted into surface meshes using surface reconstruction techniques. Among the available reconstruction methods, we primarily adopted the traditional ball-pivoting algorithm due to its robustness, computational efficiency, and ability to preserve the fine branching structures commonly found in plants. This conversion produces mesh representations that retain the geometric characteristics of the original scans while providing a closed surface required for downstream processing.

The second preprocessing task involved the generation of synthetic plant models using Lindenmayer systems (L-systems). Procedural generation enables the creation of a diverse set of plant architectures with known ground-truth skeletal structures, making it particularly valuable for training and benchmarking reconstruction models. However, because L-systems are capable of producing arbitrarily complex branching patterns, not every generated model was biologically plausible. To ensure that the synthetic dataset reasonably reflects real plant morphology, each generated mesh was manually inspected prior to inclusion, as well as the underlying mathematical rule. Particular attention was given to identifying overlapping or intersecting branches, which are physically impossible under normal plant growth, as well as branches extending below the base of the primary stem. Such configurations would correspond to branches growing into the ground rather than away from it and therefore do not represent realistic botanical structures. By filtering these unrealistic cases, the resulting synthetic dataset better approximates the structural characteristics of real-world plants while maintaining the diversity afforded by procedural generation. Moreover, these synthetic data might have a significant paucity of noise, so we manually incorporated that into the generated meshes. Ultimately, we wrote two Python packages, one that focused on branch-only mesh generation, and the other that could incorporate customizable amounts of foliage. Of course, as discussed, the real-to-sim gap would remain important to consider, so we added a parameter in both scripts to add adjustable amplitudes of surface noise to mimic that of the real-life data we used. 

Of course, the most accurate version of surface noise just from our data pipeline above is the conversion between points and surfaces. Thus, we used a custom-developed autoencoder script that could create finer or coarse closed triangular faces, converting from clean, synthetic meshes to a dense point cloud and then back. This gives us the opportunity to create the exact same irreducible error present in our observations of the GaussianPlant borrowed data. Thus, with this final step, we now had an arsenal of tools---both synthetic and extant---that allowed us to effectively handle our experimentation pipeline.

\section{Model Design and Experimentation Pipeline}
We treat the next two sections as a sort of logical narrative, displaying our thought process behind each step. Although it is unorthodox to include research failures in such a paper, we believe it is informative to understand the reason we took certain measures and precautions, as well as to give a better sense of context behind what worked and why. Hence, we outline the entire train of thought, from the initial refinement of UniRig down to the final, optimized model. As such, for the reader's purpose, we split our methodology into two parts: skeletonization and foliage-based/full-mesh rigging. The former constitutes the most important part of the project, as without a proper bone structure, attaching leaves and other appendages would be meaningless, since the underlying backbones wouldn't even be set. Thus, the bulk of the time we had allotted to our endeavors we spent on refining the topological armature of the plants. We use the remaining time to handle the rest of the distribution, the latter in our experimentation pipeline.

\subsection{Skeletonization}
The first thing to isolate was UniRig's base model's failures in providing an accurate bone layout. Without these, it would be impossible to narrow down the model's shortcomings and therefore adapt them for plant skeletal reconstruction. As such, our experimental logic begins with the use of L-systems to generate plausible yet branch-only meshes. We focused on collecting archetypal data at first, fitting some branching schemes that one might find in the real world, which at this stage included: \textit{monopodial, sympodial, bushy shrubs, vines, whorled}, and \textit{rosettes} (like palm trees' crown). The generation rules all start from variable X, and for this list are detailed as follows: monopodial trees $$X=F[+B]F[-B]X$$ $$B=FB$$ for sympodial trees $$X=F[\&+X]F[\hat{-X}]F[\&-X]F[\hat{+A}]$$ for bushy shrubs $$X=F[\&+X][\&-X][\hat{+X}][\hat{-X}]$$ for helical and step vines respectively $$\{X=F+F/X[+A]$$ $$A=F\}$$ $$\{X=F+F-X[+A]$$ $$ A=F\}$$ for whorled $$X=F/(90)[+A]/(90)[+A]/(90)[+A]/(90)[+A]X$$ $$ A=FA$$ and for rosettes (the likes of aloe plants, cycad crowns, and palm tree crowns) $$X=FFFB$$ $$B=[\&+A][\&-A][\&++A][\&--A]$$ $$ A=F$$ The other parameters, such as angle, length, and minimum/maximum iterations could be toggled to the user's preference and therefore don't matter as much.

The primary features that stood out to us in initial inference was UniRig's lack of branching and the freedom of the program to choose its own starting point. For instance, in some of the meshes that we generated, UniRig gave us a single spine down the middle, despite having clear branches veering off the central axis. Moreover, the skeletal root would sometimes start at the center of mass of the mesh, intuitive if and only if the mesh is an animate, spinal object. For examples, see the \textbf{Results-Skeletonization} section.

Thus, we tried to isolate this problem by heavily reducing the complexity of the issue at hand. Instead of a dense, tree-like structure, we chose to generate simple meshes, such as a Y-shape and a simple linear curve. From here, varying different parameters of the mesh exposed different aspects of UniRig's behavior, including the ones mentioned above. 

Once we had isolated the issues for certain, the next step was to figure out whether or not they would be fixed by one of three things: architectural changes (such as hyperparameter choices), constraint injection, or fine-tuning. The last of these options we considered last, as it is the most expensive. For the first, we varied countless parameters in the configuration files, specifically isolating the lack of branching behavior. Instead of UniRig's native \textit{VocabSwitchingLogitsProcessor} class, we developed our own processor classes \textit{BranchIsolationLogitsProcessor} and \textit{BranchBoostLogitsProcessor} to use as a substitute instead. The former broke down the behavior of the autoregression during the \textit{tokenization} step, outlining whether or not branch tokens were even emitted in the process of creating the token sequence. The way that UniRig works is that it first generates a sequence of tokens that correspond to positions of joints, as well as specialty tokens such as a beginning-of-sequence (\textit{bos}) and end-of-sequence (\textit{eos}) token. This sequence is then detokenized during inference, generating a hierarchical armature as the result. One of these special examples would be the branch token, corresponding to number 256 if observed in the token lineage; if and only if this is present, then will there be any branching at all. This gave us a sense as to whether or not UniRig's learned prior was influencing its output. We quantify our results and show some examples in the later sections. BranchBoostLogistProcessor took the logits score for the branching token and boosted it by a set amount at each quantization step, allowing for it to climb the ranks and potentially go into sampling for the next token to use in the output sequence. The thought was that the learned prior could be stifling the choice of the branch token, so boosting its likelihood could potentially fix the lack-of-branching issue. 

Another approach that we took to help adapt the rigging performance to vegetation was to enforce a starting point, or root. As stated earlier, the autoregressive algorithm had free choice as to where to place the first joint node---and thus the starting token, excluding the \textit{beginning-of-sequence} token---which is counterintuitive for plant-based meshes. The reason for this is that plant growth behaviors exhibit a supervised growth, usually starting from a single point and then slowly extending out in branches or trunk. Thus, the theory was that forcing a root node would allow for the program to naturally follow a plant's growth patterns. To implement this, we needed to inject code that correctly tokenized a constrained position of this growth origin. A naive but practical approach we used was to find the vertex of an upright mesh with the most negative (least) z-coordinate, which is the "up" direction in Euclidean mathematics. Then, we employ UniRig's internal positional encoding arithmetic: \begin{equation}
    M: x \in [-1,1] \mapsto d = \left\lfloor \frac{x+1}{2} \times D \right\rfloor \in \mathbb{Z}_D
    \label{eq:unirig-tokenize}
\end{equation}

\begin{equation}
    M^{-1}: d \in \mathbb{Z}_D \mapsto x = \frac{2d}{D} - 1 \in [-1,1]
    \label{eq:unirig-detokenize}
\end{equation} 
\noindent where
\begin{itemize}
    \item $d$ is the discrete token recovered from the sequence (an integer bin index produced by the autoregressive model)
    \item $D = 256$ is the fixed number of bins used to discretize each axis
    \item $\frac{2d}{D}$ rescales $d$ from $[0, D)$ back to $[0, 2)$
    \item subtracting $1$ shifts this to the original normalized range $[-1, 1)$,
    \item $x$ is the reconstructed continuous coordinate, an approximation of the original normalized joint coordinate along one axis
\end{itemize}

The last resort was simply to continue training the model off of accurate plant meshes, or finetuning the model itself. We wrote custom Python programs to procedurally generate archetypal, branch-only data to use, encompassing a large variety of existing genuses of plants. For this step, we included the six discussed above as well as groups such as cycads, columnars (like a cactus), rosettes, and rhizomatous plants for a total of 12 different ones. Later, we decided to omit rhizomatous plants, as stem's behavior---in being underground---introduced too many new variables for too few data points. As such, our updated custom script generated branching meshes and their ground truth skeleton, and we passed them into UniRig for finetuning purposes. Thus, we now had an aggregate script to generate a uniform number of a total of eleven different families: monopodials, sympodials, vines, bushy shrubs, cycads, palms, whorleds, basal rosettes, weepings, columnars, and round bushes. This gave us enough data for training that was representative of real-world species, and we generated fifteen thousand samples spread evenly, yielding about 1364 samples per group. 

After this run of fine-tuning, we engaged in another round of evaluation for the model. This time, however, to help bridge the real-to-sim gap, we needed a way of providing some reasonable noise to the near-perfect archetypal meshes that we already could generate. Looking towards our sister project, GaussianPlant, the primary source of noise from those meshes that we used was surface reconstruction noise such as in Figure 2, meaning that the mesh surface was highly uneven. While this was true, any branching patterns were still easily captured. Thus, we introduced a new variable parameter into our branch-only mesh generation script, allowing for the user to toggle the amount of surface noise that they wanted to introduce to the model. Using these noisy synthetic data, we then compared the finetuned model's performance with that of a perfect procedural mesh. 

The last thing in the primary component of our work was to see if this generalized to our synthetic data. However, instead of only forcing a certain amount of uninterpretable noise into our mesh data as with natively in our script, we opted to use an autoencoder program, which we wrote ourself. We took in meshes generated from the aforementioned branch-only script, spreading across eleven archetypes, and then passed them through our autoencoder. What it does is simulate real measurement noise, rather than injected noise. The input mesh would first be converted to a dense point cloud, much like that we found in GaussianPlant's data, and with a certain number of points ranging between 100 thousand and 500 thousand. Then, we apply the classic ball-pivoting algorithm, since it is usually less expensive for a less dense cloud and preserves a higher level of detail in sealing the mesh, even if the result has some holes or discontinuities. We did try using Poisson surface reconstruction, but the tendency to produce smooth surfaces meant too much noise in the output that was incomprehensible to even the naked eye. Thus, after the autoencoded process was done, we were left with a mesh that had surface noise akin to that of metric error, a systematic error that is present in all scanning techniques. We built this regime with the real-to-sim gap in mind, ultimately proving to be the most consistent to real life data. With all of our testing data---including GaussianPlant's real plant meshes from existing plant species scans---we evaluated our finetuned model, the results of which can be found below in the \textbf{Results-Skeletonization} section.

\subsection{Full Mesh - Leaves and Lesser Appendages}
The next thing to tackle was the issue of leaves, the only other part of a plant that was significant. While a short one, this actually is quite a complicated problem, as leaves have far more complex "weights" for movement than that of articulated characters. For instance, just consider the differences between pine needles, palm fronds, grass blades, and maple leaves. They all have distinct internal features and components that make up their non-skeletal aspects.

Thus, we first start by testing out the baseline segmentation of the leaves with the existing empty \texttt{<type>} token that's built into UniRig's implementation. To do this, we needed a new aggregate generation script that accounted for leaves and minor appendages. Thus, we imported the same \textit{archetypes.py} file that we had previously written for the branch-only meshes to preserve the branching behavior, while incorporating it into a completely new data package that added a large variety of functionality. We needed a robust way to create proper synthetic data, so we added customizable attributes such as leaf thickness and foliage density, since a sympodial tree like an oak tree might have very thin leaves while the same shape might take on a thicker representation in the eyes of fruit trees like mulberry. In other words, we developed another synthetic generation package, \textit{leafy\_gen}. Furthermore, for the sake of the real-to-sim gap, we again incorporated surface noise to the meshes, this time including the leaves and petiole as well. We use the two-dimensional Gaussian splatting meshes from GaussianPlant to do more testing, along with these synthetic meshes.

The intial results were quite excellent but confusing, as it seems that the finetuned model performs well on the now-leafy plants, although the leaves were stripped from the branches. This ended up spitting back output meshes that only had the branches, which could be problematic since it implied that this layer of testing was no different than that of without leaves. As such, we observed ways in which our current mesh was attaching the leaves to the branches themselves. Originally, our script generated the branches first, then adding leaves as an appendage connecting point. However, it could be that due to both a learned prior as well as the fact that the branching mesh was already a closed surface, that the leaves were absent during runtime. 

Hence, we overhauled the petioles to perform similarly to the branch tube meshes, simply making them like thin tapering branches using the same logic, and then linking to a leaf at the tip. This way, the entire mesh is hermetic and airtight, without any holes or obstructing surfaces between appendages. This time, under inference, the leaves stayed attached and actually compiled together with the skeleton. However, due to this new framework in UniRig's observed input, some of the bones shot into and past the leaves, treating them as part of the branches. Thus, it was imperative that we run a different checkpoint that actually incorporates the leaves as part of the mesh, a slightly different but highly varied input space compared to the branch-only structures from the last trained checkpoint. 

Due to this, we ran a final finetuning run, generating 17,600 samples across the same eleven archetypes, the most common found throughout the world. Of course, this means that we had 1600 total samples for each archetype respectively, giving us a highly diversified and stochastic yet realistic mesh dataset for training. Once we finished this final round of training, the step in our evaluation was running inference on test data, particularly the autoencoded plant meshes using our custom autoencoding script. Unfortunately, most of the real plant data that we had access to were branch-only, since they were taken from 3D Gaussian splatting experiments from previous years. The results of our final checkpoint and results are below, and we provide visualizations of our full meshes there too, to avoid cluttering the narrative here.

\section{Results}
\fontsize{11}{13}\selectfont
We now take the narrative we presented in methodology and outline the results that we achieved along the way. Again, it is important to note that while heterodox, it is equally important to understand the shortcomings of our strategies in order to better motivate the later steps that we took. Hence, the authors present the same chronology in logic corresponding to the same subsections above, limning the important results that came out of each phase of the project. 

\subsection{Skeletonization}
The opening scheme of our analysis involved the use of Lindenmayer systems to generate archetypal, foliage-less meshes, particularly for sparking our understanding of why the base model has certain deficiencies. We show some examples now. As noted previously, the base model of UniRig experienced highly linear rigging behavior, with a lack of branching at all. We can see this behavior in Figure 3, which shows a mesh generated using a Lindenmayer system that fits the whorled archetype, resembling a mix between a typical monopodial tree and a traditional pine tree. We see here that the original UniRig program tends to lean towards the articulated characters, or those with a clear and defined spine. For instance, with humanoid meshes, one would expect that the skeleton has a primary linear segment through the center of the mesh. We see that behavior exhibited clearly in this case, among many others.
\begin{figure} [h]
    \centering
    \includegraphics[width=0.7\linewidth]{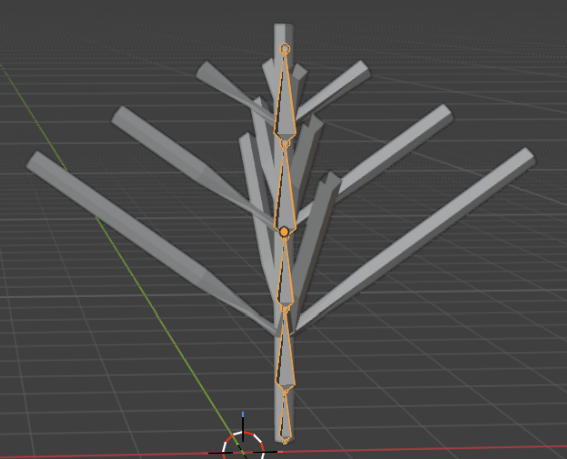}
    \caption{Whorled archetype, initial UniRig inference}
    \label{fig:placeholder}
\end{figure}
\begin{figure} [h]
    \centering
    \includegraphics[width=0.7\linewidth]{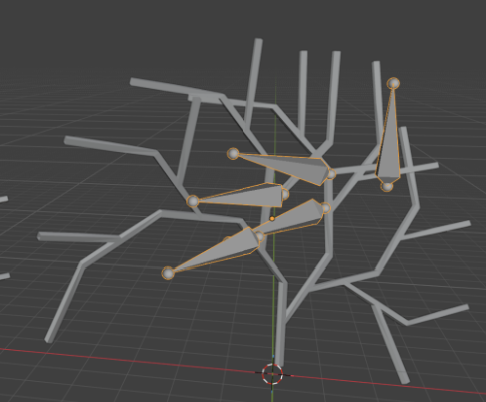}
    \caption{Tree-like L-system mesh, initial UniRig inference}
    \label{fig:placeholder}
\end{figure}
Moreover, it also had complete freedom to place the root---or starting point---of the skeleton wherever it wanted. For instance, the branching point for a human might be the sternum, or the center of the chest, since the limbs and spine all spread out from that point. We note that without loss of generality, this could be labeled as the "center of mass" depending on the character. This concept distinctly appears in Figure 4, where the root is placed in the heart of the sympodial tree, since the joints all branch out from an arbitrary position in the mesh. The reason why the whorled example in Figure 3 doesn't look like this is likely from the fact that in the model's point of view, the mesh resembles a spinal figure with plenty of tiny, negligible limbs. if you rotate the model $180$ degrees, then we see that it looks like the skeleton of a tubular fish, explaining the spinal behavior. 

To better see these issues in action, we heavily reduced the complexity of our input space. Simpler mesh shapes didn't provide promising conclusions either. We tested a twig-like, symmetric Y-shape with a few varying characteristics such as angle between the fork, length dilation, and overlapped branches, but the results all fell into the following two cases. For starters, with wider angles between the branches of the Y vertically, the same free-root estimation issue persisted, as in Figure 5. 
\begin{figure} [h]
    \centering
    \includegraphics[width=0.6\linewidth]{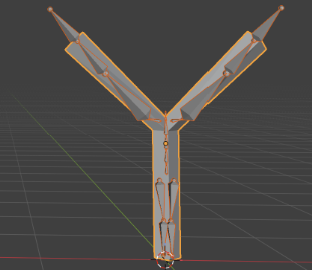}
    \caption{Y-shape mesh with wider angle, initial UniRig results; note the position of the root joint in the center of the mesh}
    \label{fig:placeholder}
\end{figure}
\begin{figure} [h]
    \centering
    \includegraphics[width=0.5\linewidth]{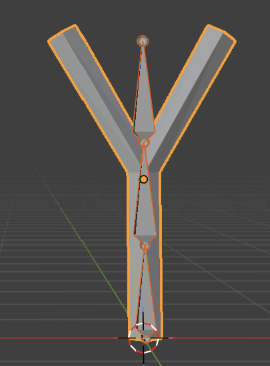}
    \caption{Y-shape mesh with tighter angle, initial UniRig results; primary spine that encompasses the volumetric height of mesh as in Figure 3}
    \label{fig:placeholder}
\end{figure}
We can see that while the entire mesh is accounted for by the skeleton, the root node occurs at the branching point in the middle of the object itself, a parallel result to earlier with the sympodial tree in Figure 4. Moreover, in Figure 6, we note that the spinal prior remains, even in this reduced case.  

The first problem we decided to try and tackle was the lack of branching, including the spinal prior. The best case scenario would be if we could solve that issue without needing an entire finetuning run. As such, we first examine the inner workings of how branching behavior is exhibited during runtime in the autoregression. As outlined above, UniRig determines the existence of such behavior during the tokenization step, when it generates the sequence of positional and indicator tokens. Thus, to visualize this process, we developed a custom logits processor class called \textit{BranchIsolationLogitsProcessor} that preserved the logic of UniRig's processor, but injected sampling outputs throughout the sequence formalization. To reiterate, in the sequence of tokens that are emitted for the skeleton topology to follow, if the branch token (256) is absent, then there wouldn't be any branching behavior to follow. Thus, running this on the Y-shaped mesh found in Figure 6, we have the result in Figure 7, which clearly indicates that there isn't any branching happening internally. 
\begin{figure} [h]
    \centering
    \includegraphics[width=1\linewidth]{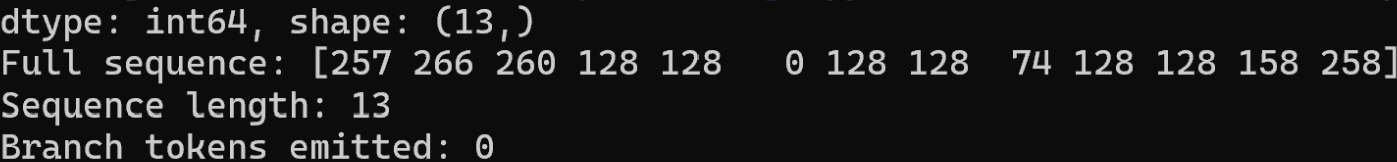}
    \caption{Y-shaped mesh token sequence, UniRig base model inference with zero branch tokens emitted}
    \label{fig:placeholder}
\end{figure}
The same outcome appeared for the archetypal branch-only meshes that we initially tried on first inference. Thus, the next in the pipeline was trying to boost the logits score of the branching token to see if the internal token sampling would actually end up choosing it and thus exhibiting proper plant behavior. We developed another custom logits processor class, \textit{BranchBoost} that kept the same properties of the base autoregression, but dilated the logits value by a user-specified amount. The results were both promising and discouraging at the same time. For a very specific angle arrangement and property selection, we found that boosting the score worked great (see Figure 8). \begin{figure} [h]
    \centering
    \includegraphics[width=0.5\linewidth]{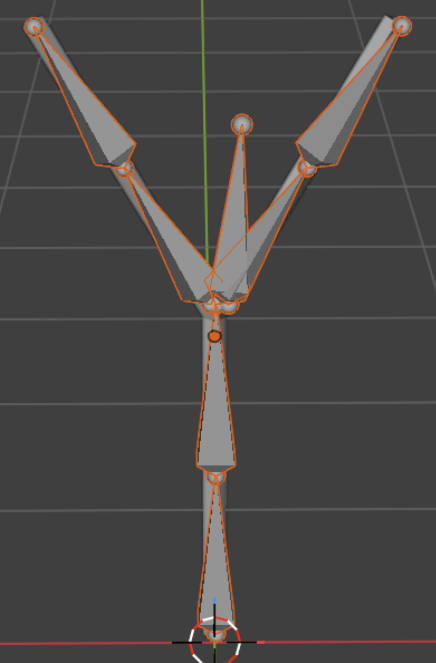}
    \caption{Y-shaped mesh with branch boost, angle between spokes at 60 degrees}
    \label{fig:placeholder}
\end{figure}
However, for most of the other meshes, the result was a complete breakdown of the regime, since boosting the branch token in tokenization resulted in an out-of-distribution set, with highly degenerate cases as in Figure 9. \begin{figure} [h]
    \centering
    \includegraphics[width=0.5\linewidth]{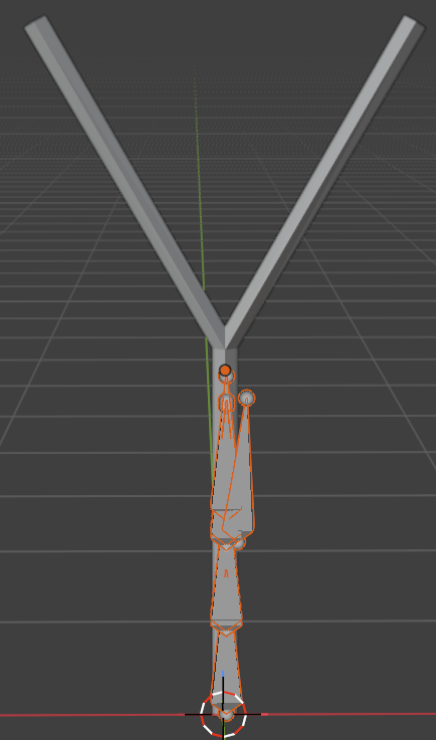}
    \caption{Y-shaped mesh with branch boost, same as in Figure 8 but dilated by a factor of 2}
    \label{fig:placeholder}
\end{figure} 
With this, one road came to a dead end.

The second issue from above was the freedom of choice in the position of the root node. This could be another reason why potential branching behavior was suppressed as well, since the model might see a specific alignment of the mesh as the "spine." To fix this, we directly injected code into UniRig's tokenization and detokenization process to make sure that the root node position was specified by the user during runtime, following the logic shown earlier. What ended up happening was largely out-of-distribution again. For example, when forcing the root to be at the correct location, with one of the most simple branches \textit{without branching at all}, we find what we see in Figure 10. \begin{figure} [h]
    \centering
    \includegraphics[width=0.65\linewidth]{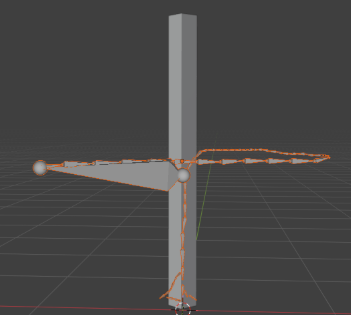}
    \caption{Simple line mesh, with initial root-forcing test}
    \label{fig:placeholder}
\end{figure}
\begin{figure} [h]
    \centering
    \includegraphics[width=0.65\linewidth]{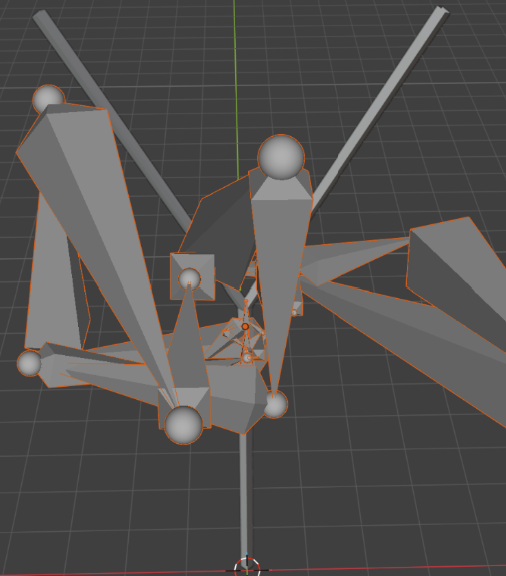}
    \caption{Y-shape branch mesh, with initial root-forcing test}
    \label{fig:placeholder}
\end{figure}
With Figure 11, we find a different story, since this time the mesh already has branching. However, the model now spits out an incredibly debauched skeleton that has no interpetability whatsoever. What might be happening is that when we introduce another constraint to the tokenization/detokenization process, we are significantly increasing the dimensions of the already-complex input space. Then, due to the limited learned prior of UniRig, we find that these iterations of inference run outside of the learned distribution, thus providing us with these chaotic rigs. Clearly, these two direct solutions did not solve our original problems at all. It is worth emphasizing that we did not end up using both branch boosting \textit{and} root position forcing at the same time, since both of these were highly degenerate on their own. In other words, it isn't a stretch to say that using both strategies in tandem might crash the model intrinsically, although the reader is encouraged to reproduce these on their own.

Due to this, we eventually turned to fine-tuning in multiple rounds. First, we wrote a Python script that procedurally generates a tree-like mesh with only sympodial branching, alongside its ground truth skeleton. After the training process converged, we noted that this might not represent nearly enough genuses of plants for our result to be useful, although it might provide insight as to the applications of the autoregression. This is highly evident in Figures 12 and 13, which are synthetic meshes of both a whorled tree and basal rosette, the latter being akin to an aloe plant. \begin{figure} [h]
    \centering
    \includegraphics[width=0.6\linewidth]{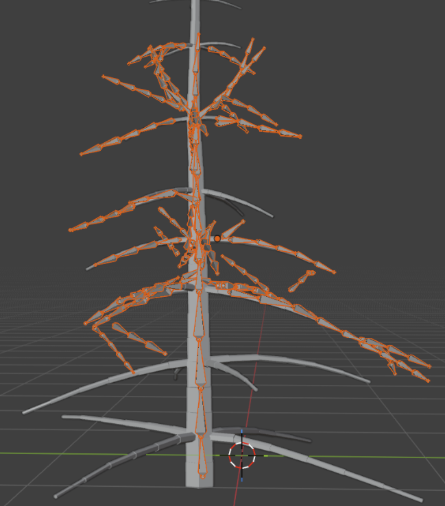}
    \caption{Whorled tree synthetic mesh with predicted armature, after sympodial-only finetuning process}
    \label{fig:placeholder}
\end{figure} 
\begin{figure} [h]
    \centering
    \includegraphics[width=0.6\linewidth]{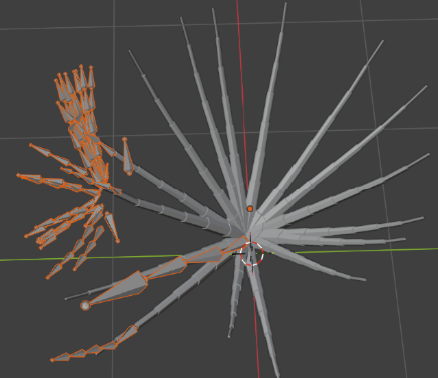}
    \caption{Basal rosette synthetic mesh with predicted armature, after sympodial-only finetuning process}
    \label{fig:placeholder}
\end{figure}
To fix this, our second Python script generated archetypal training data off of eleven different archetypes, which are listed above, including those in the aforementioned figures. With this newfound script, we not only could represent a wider selection of plausible plants, but as mentioned earlier, one could toggle the amount of surface noise that was present in the mesh generation, allowing for a more realistic form of data that didn't corrupt the program as is. With this arsenal of practical synthetic meshes, we engaged in a longer, more thorough round of training with 15,000 data points. 

Not only did this solve the lack of branching behavior present with the original UniRig model, it was now also possible to properly teach a position in the root node, encouraging the predicted skeleton to mimic the growth patterns behind real-life vegetation. With our test meshes, this new model performed near perfectly. In Figures 14 and 15, we have two different archetypes that we considered, cycads and monopodials. In the same image, we have their clean and noisy meshes, along with their inference results from left to right respectively. The angle we viewed them at was slightly different when capturing the model for viewing purposes, but both meshes are the exact same generation seed, just with or without noise. We find that after our prospective training runs, the model now performs excellently in terms of the skeletonization process, evidence of the natural capabilities of autoregression for our purpose. We assert that autoregressive models are particularly well suited for plant skeletal reconstruction because plant structures are inherently hierarchical and recursive. Starting from the primary stem, each branch gives rise to children in a sequential relationship, closely resembling the token-by-token generation process employed by UniRig. It also makes sense due to the same parallel logic that govern Lindenmayer systems and our mesh generation scripts. 
\begin{figure} [h]
    \centering
    \includegraphics[width=0.9\linewidth]{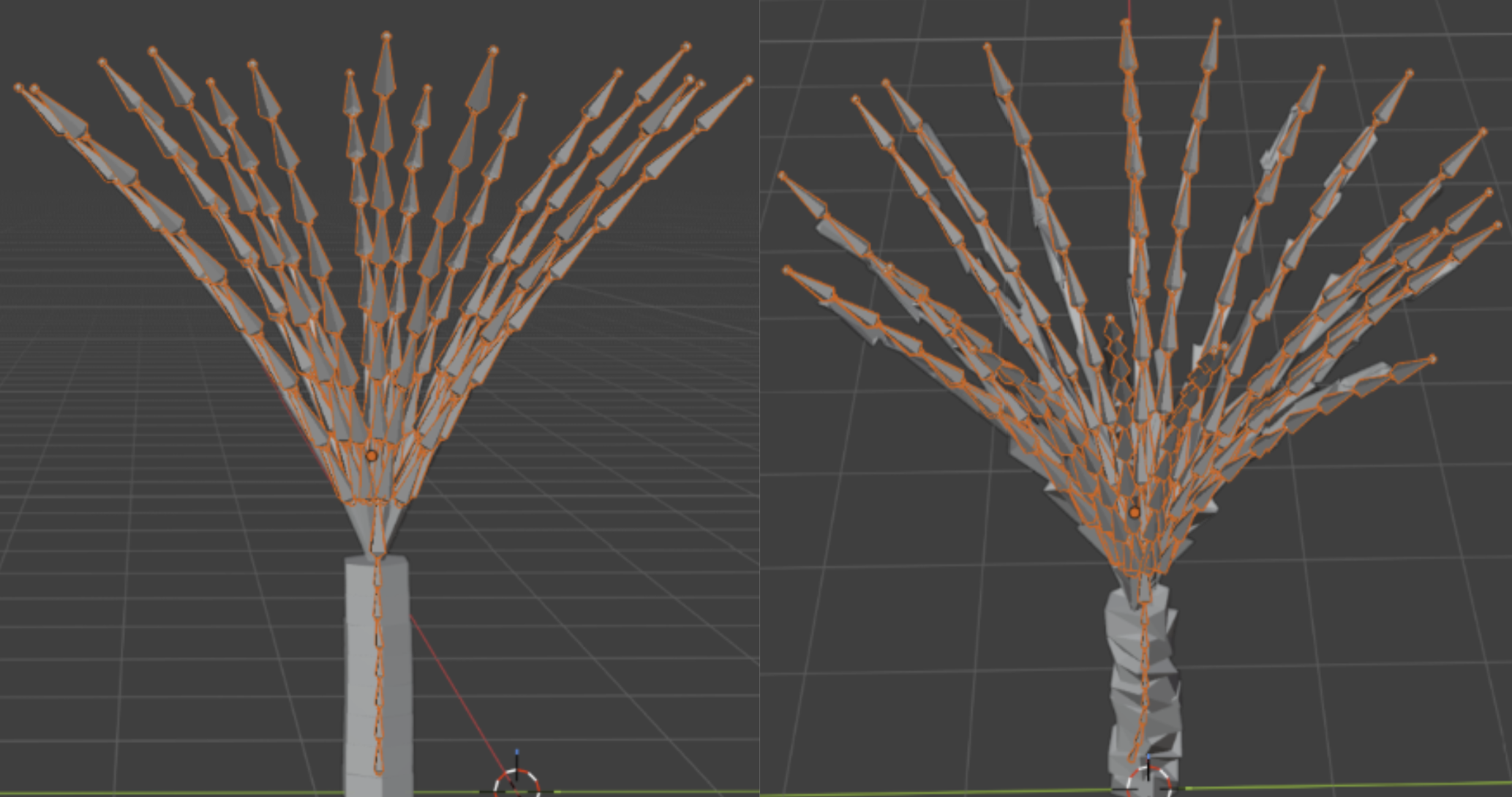}
    \caption{Cycad branch-only clean mesh (left) and noisy surface mesh (right) along with their respective predicted skeletons}
    \label{fig:placeholder}
\end{figure}
\begin{figure} [h]
    \centering
    \includegraphics[width=0.9\linewidth]{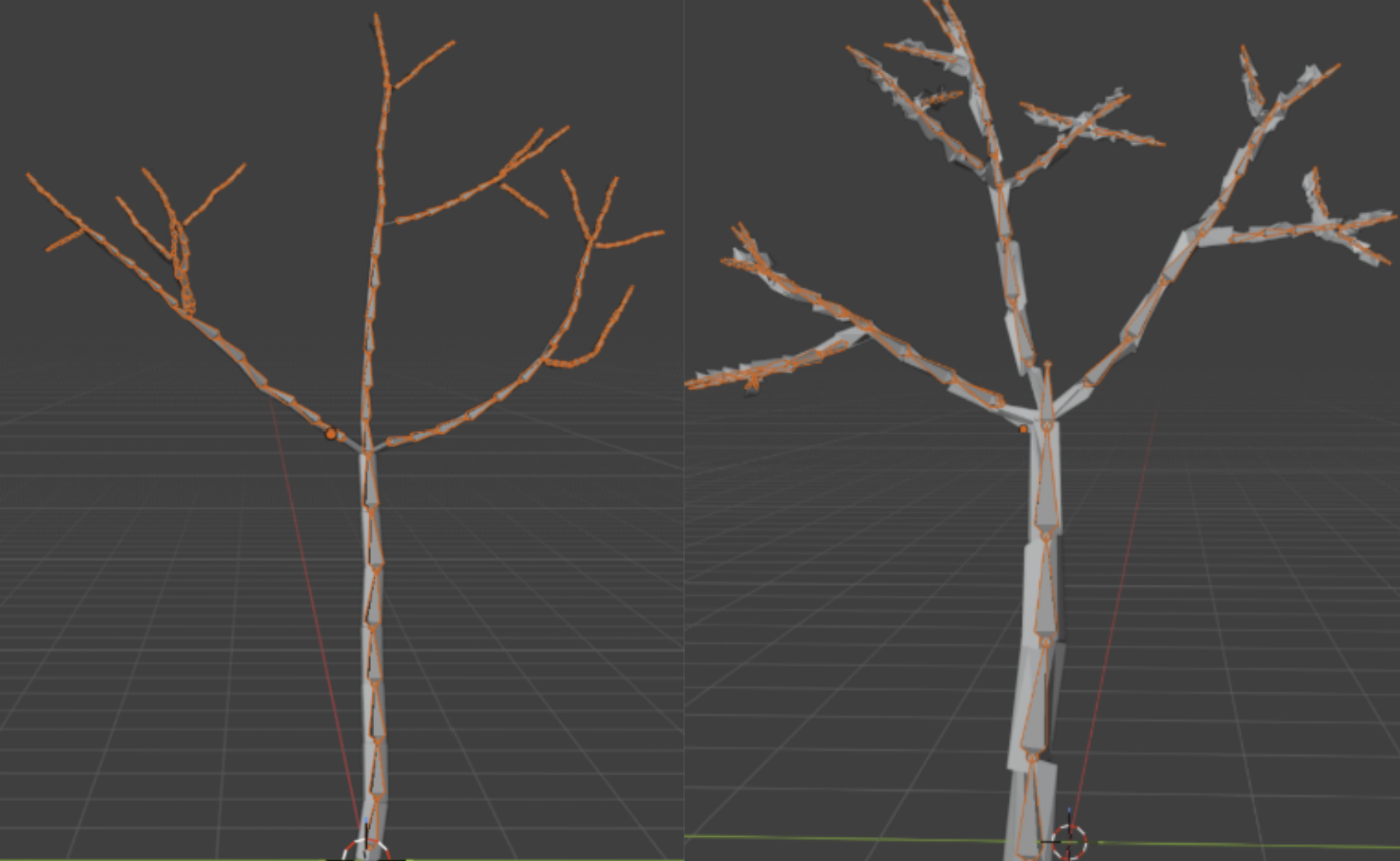}
    \caption{Monopodial branch-only clean mesh (left) and noisy surface mesh (right) with respective predicted skeletons}
    \label{fig:placeholder}
\end{figure}
Revisiting a more complex model, the same whorled test subject yields the output in Figure 16, a much more promising skeleton than before. While we still note some amount of inaccuracy in the predicted model, we see now that the program generalizes well to different types of plants, not just sympodial trees. Again, these conclusions branch off of the natural intuition that autoregressive tokenization directly imitates the growth patterns that most plants experience, starting from a single root at ground level. Even with practical amounts of noise, the skeletal structure remains realistic. In short, we now had a highly accurate skeleton for both clean and noisy synthetic data, capable of generalizing to real life data in the next phase.
\begin{figure} [h]
    \centering
    \includegraphics[width=0.9\linewidth]{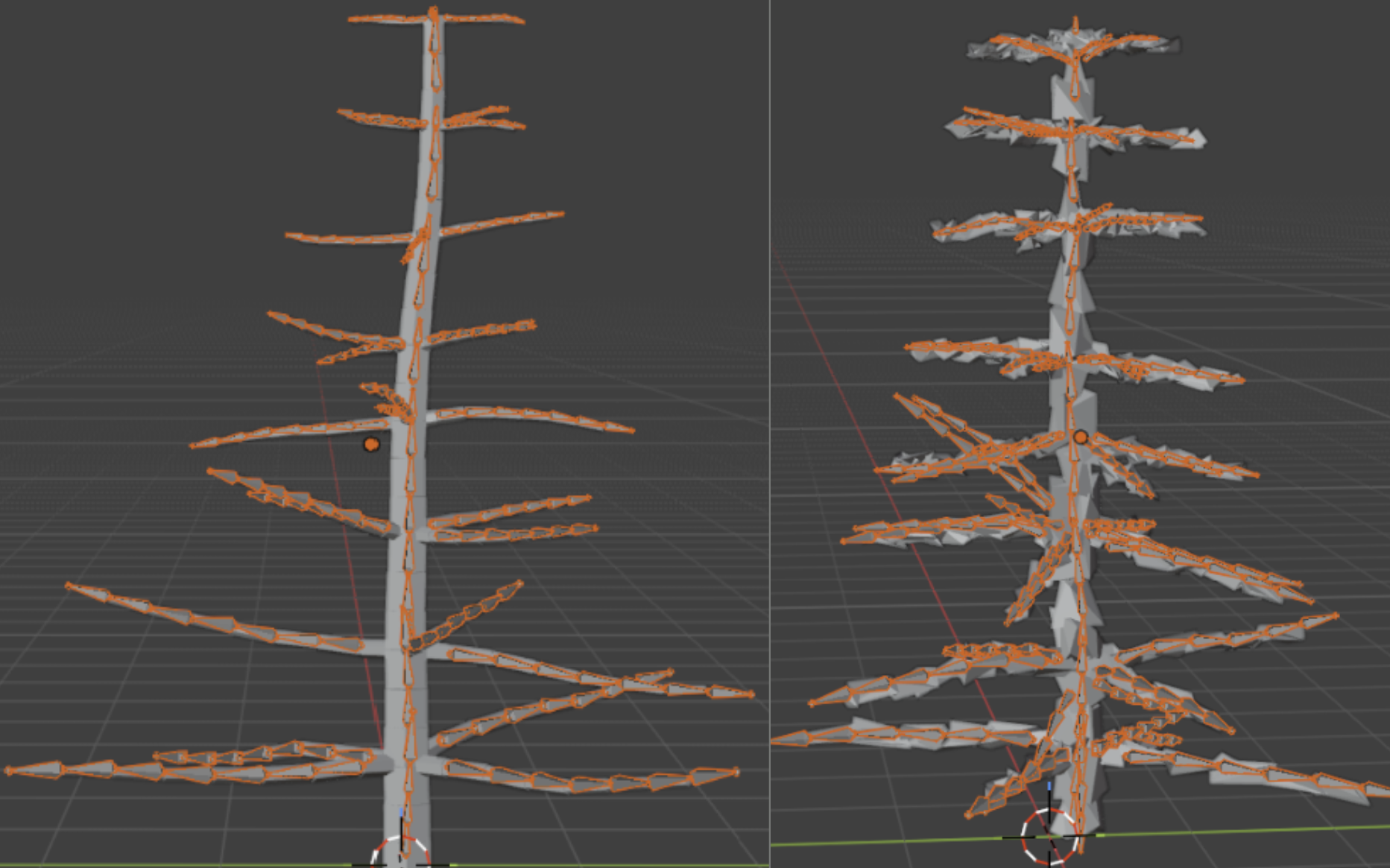}
    \caption{Whorled branch-only clean mesh (left) and noisy surface meshA (right) along with predicted skeletons}
    \label{fig:placeholder}
\end{figure}

If this wasn't convincing enough, we finish this section with two important groups of testing. First, we find that the finetuned branch-only model generalizes well even to \textit{real plants}, in data that we borrowed from GaussianPlant. As we can see in Figures 17 and 18, which feature real-life plants from past experimentation, the program generalizes well past synthetic data alone. Although there are some errors, the lack of certain branches being captured can largely be attributed to the lack of true connectivity due to measurement constraints. However, one can at least note that we achieve about 90 percent accuracy against the original loss. 
\begin{figure} [h]
    \centering
    \includegraphics[width=0.6\linewidth]{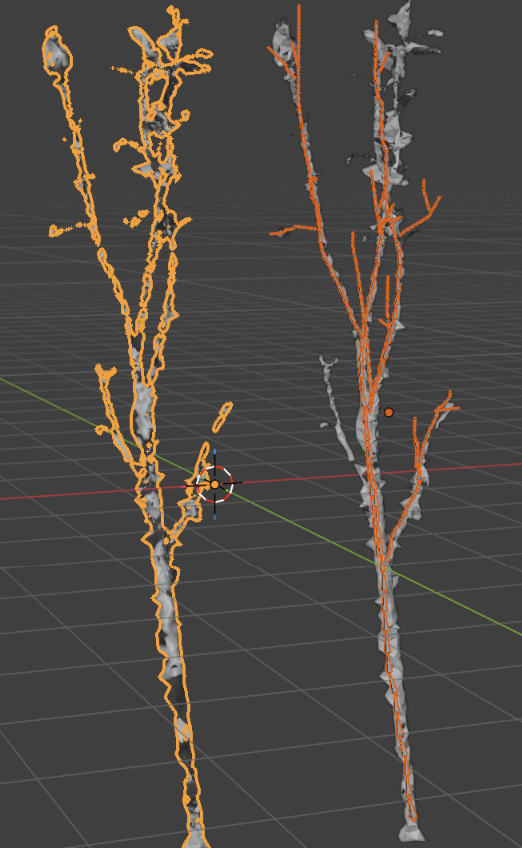}
    \caption{Real lavender plant structure, dense point cloud to mesh, final branch-only checkpoint}
    \label{fig:placeholder}
\end{figure} \begin{figure} [h]
    \centering
    \includegraphics[width=0.6\linewidth]{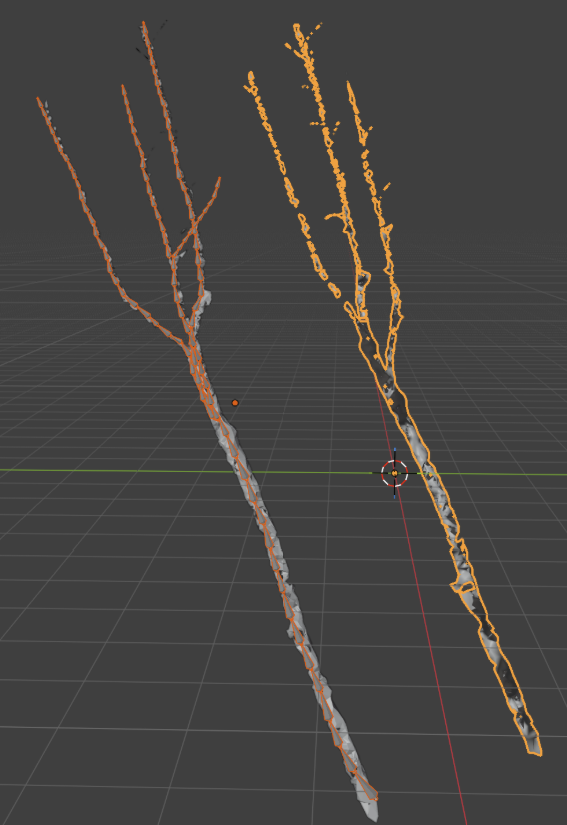}
    \caption{Real mesh of a twig/branching structure from a tree, final branch-only checkpoint}
    \label{fig:placeholder}
\end{figure} \begin{figure} [h]
    \centering
    \includegraphics[width=0.8\linewidth]{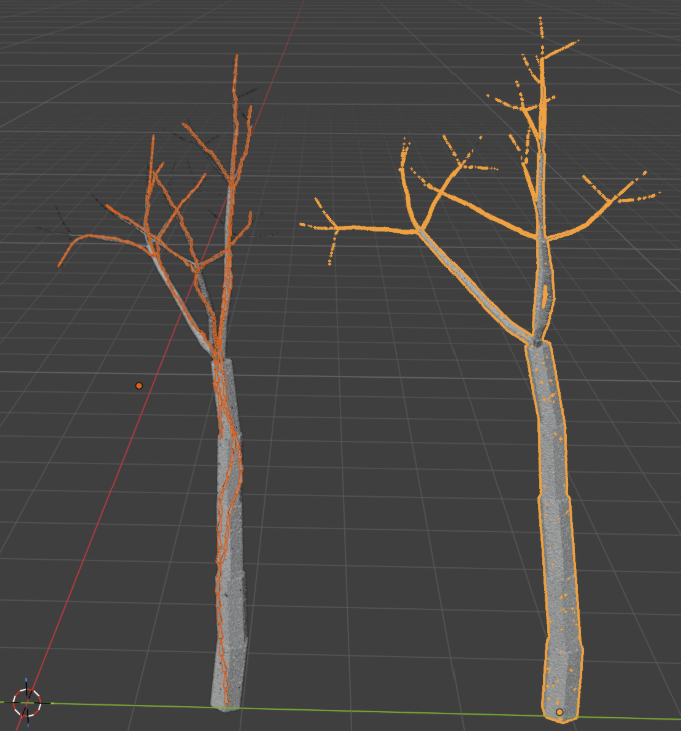}
    \caption{Synthetically autoencoded mesh: monopodial, branch-only tree and its skeleton}
    \label{fig:placeholder}
\end{figure} \begin{figure} [h]
    \centering
    \includegraphics[width=0.8\linewidth]{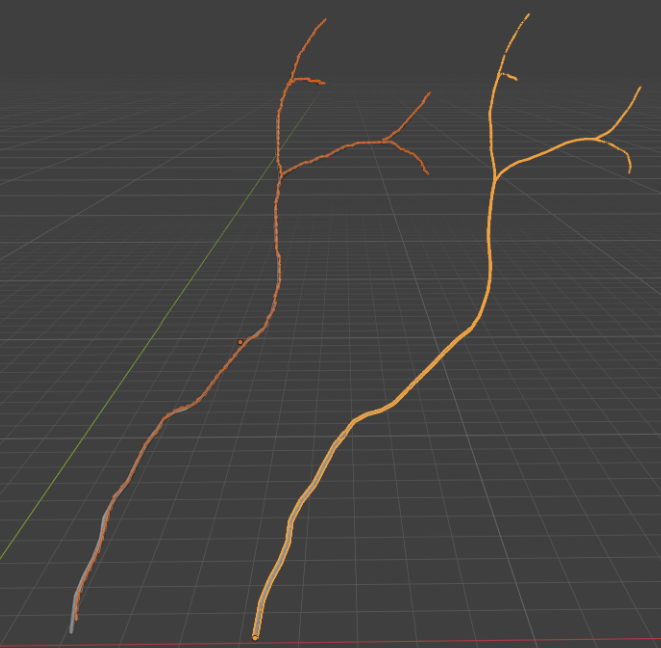}
    \caption{Helical vine synthetic, autoencoded mesh with surface noise}
    \label{fig:placeholder}
\end{figure} \begin{figure} [h]
    \centering
    \includegraphics[width=1\linewidth]{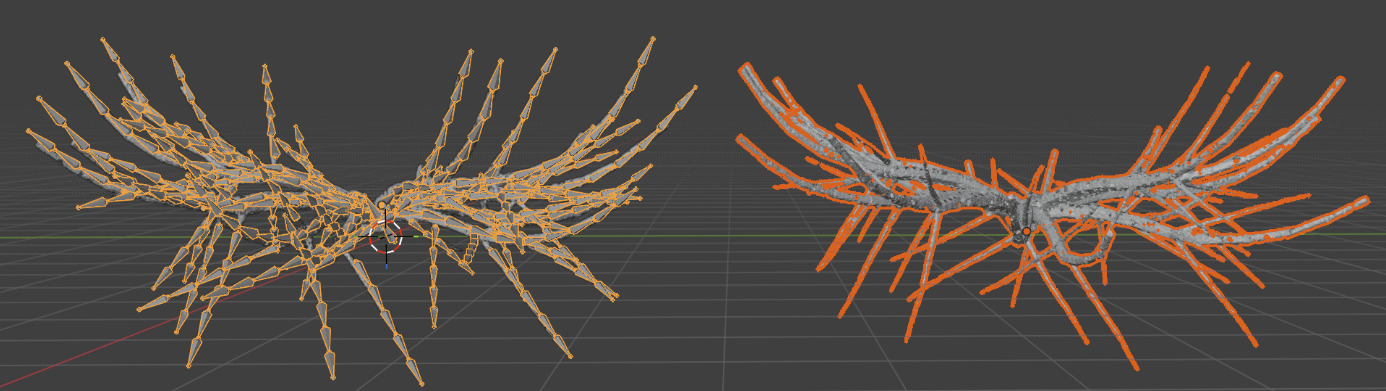}
    \caption{Complex bushy shrub (branches only), with autoencoded surface noise}
    \label{fig:placeholder}
\end{figure}

The results of using our autoencoder script compounds the successes more. After feeding our test data into it, we recover mesh reconstructions from ball-pivoting, in natural propinquity to GaussianPlant's available metrics. In Figures 19, 20, and 21, we find the results of our finetuned model from autoencoded versions of a monopodial tree, a helical vine, and a bushy shrub respectively. Even for complex, dense branching models such as the shrub, the autoregressive rigging does an excellent job, proving the natural application of UniRig's internal machinery for vegetation.

\subsection{Full Mesh Rigging}
\begin{figure} [h]
    \centering
    \includegraphics[width=1\linewidth]{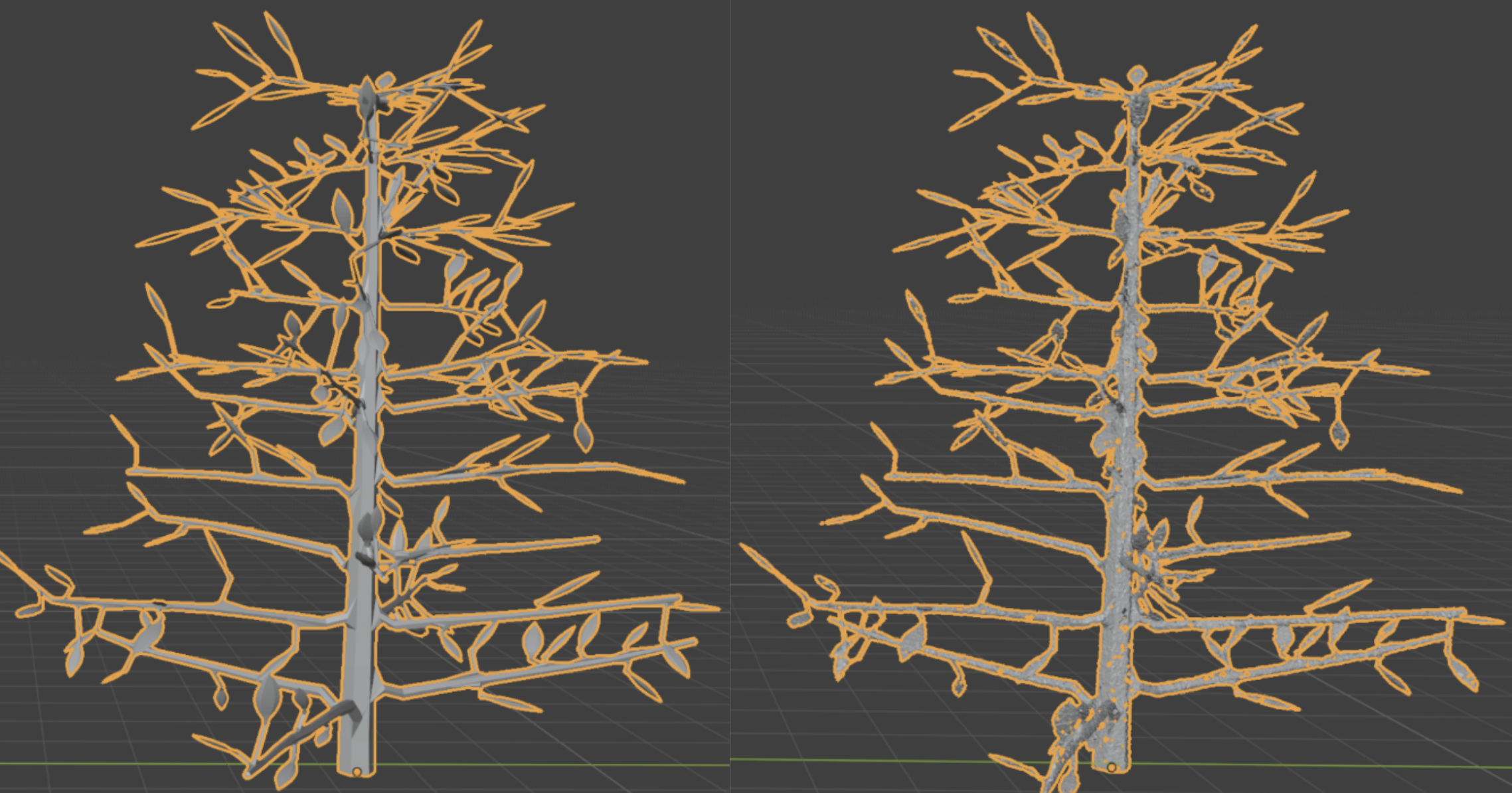}
    \caption{Synthetically generated mesh of a whorled tree, such as a redwood - \textit{(left)} original, held-out mesh \textit{(right)} same mesh, but passed through autoencoder with 150000 sample points for point cloud conversion}
    \label{fig:placeholder}
\end{figure}\begin{figure} [h]
    \centering
    \includegraphics[width=0.9\linewidth]{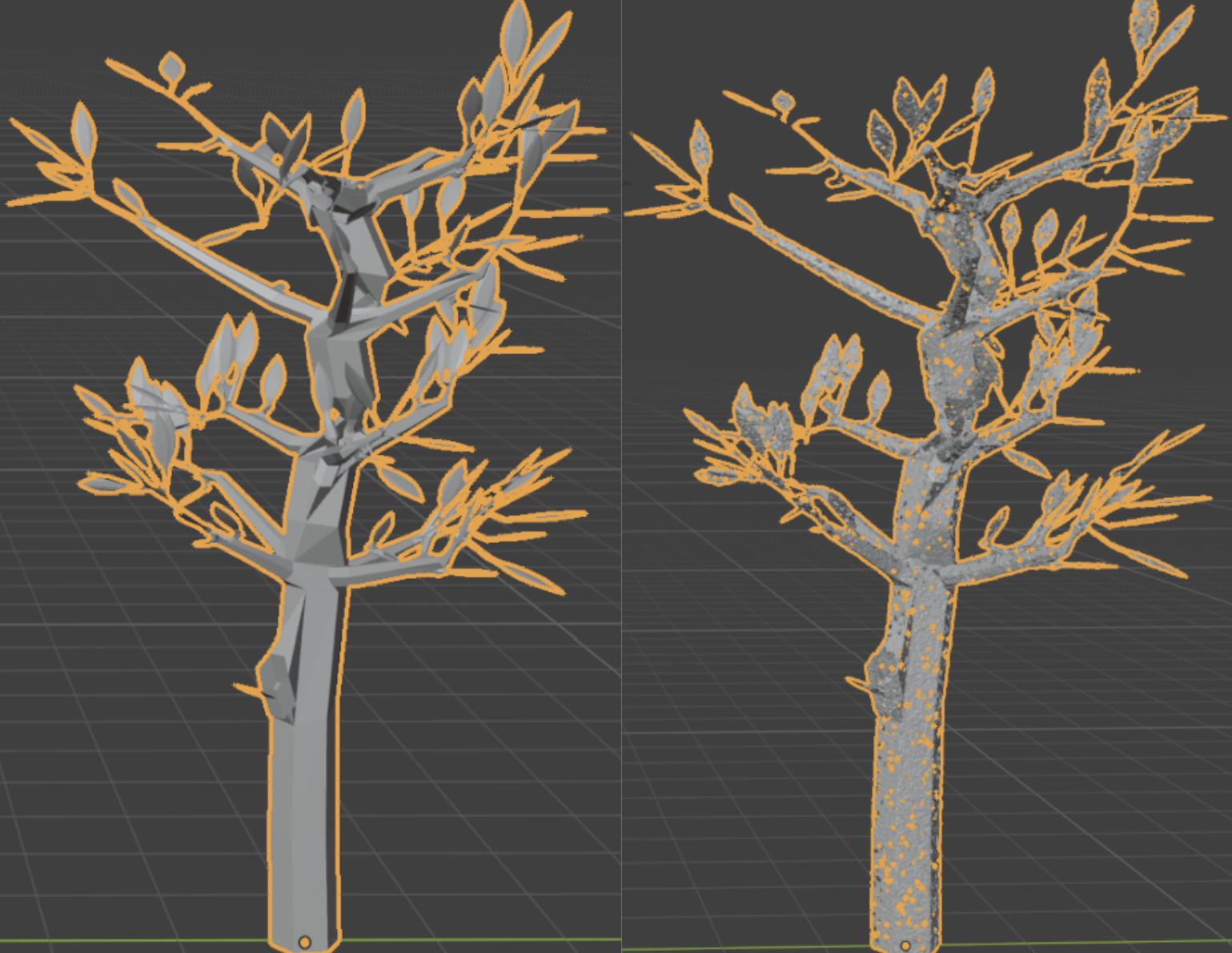}
    \caption{Synthetically generated mesh of monopodial tree - \textit{(left)} original, held-out mesh \textit{(right)} same mesh, but passed through autoencoder with 150000 sample points for point cloud conversion}
    \label{fig:placeholder}
\end{figure}
First, we start by displaying some examples of the package-generated plant meshes that had custom attached leaves, such as in Figure 22 and 23. These feature a whorled tree and monopodial tree respectively, with elliptical yet tapering leaves. They represent the final form of our \textit{leafy\_gen}, providing a glimpse into out most robust and realistic version of mesh generation out of our countless prototypes. Note that for the autoencoded mesh on the right, we see surface noise akin to measurement noise from traditional mesh capture tools, such as LiDAR scanning, among others. We see the holes created from lack of sampled points, and the ball-pivoting algorithm preserves detail down to the level of leaves and petioles as well. 

\begin{figure} [h]
    \centering
    \includegraphics[width=1\linewidth]{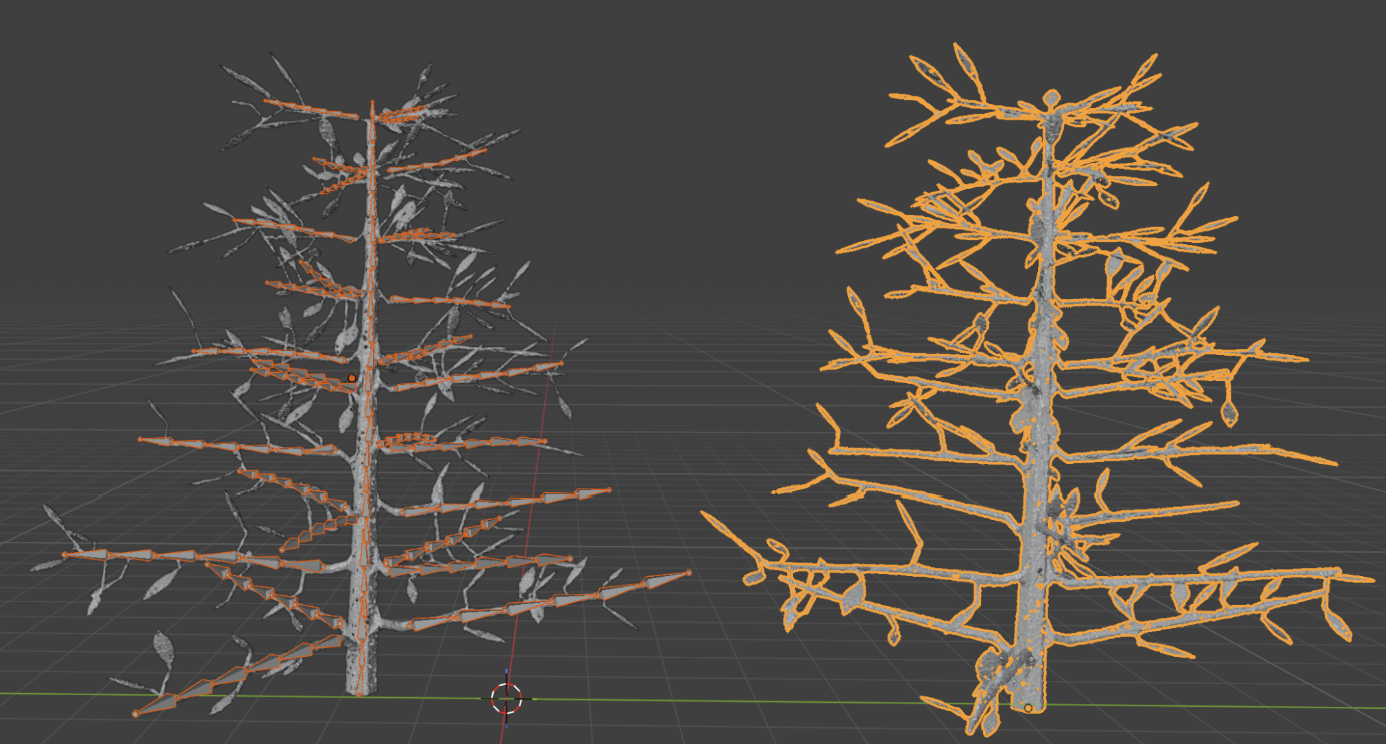}
    \caption{Autoencoded whorled tree (same as in Figure 22, \textit{right}), along with predicted skeleton \textit{(left)}}
    \label{fig:placeholder}
\end{figure}
\begin{figure} [h]
    \centering
    \includegraphics[width=0.8\linewidth]{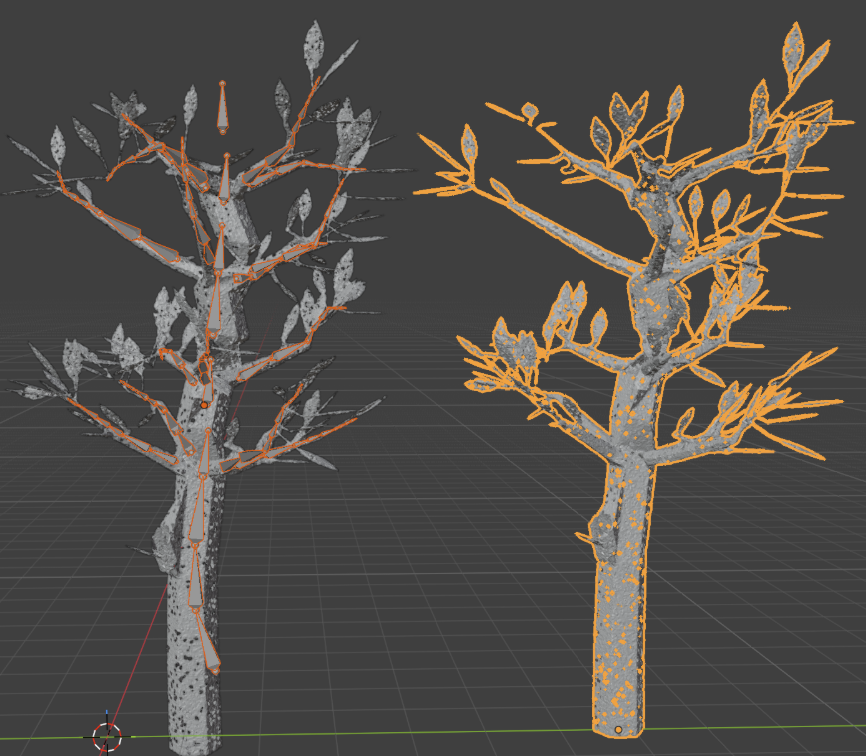}
    \caption{Autoencoded monopodial tree (same as in Figure 23, \textit{right}), along with predicted skeleton \textit{(left)}}
    \label{fig:placeholder}
\end{figure}

For better visualization of our final results after the leafy round of finetuning, we use the same meshes as in Figures 22 and 23. The results can be found in both Figures 24 and 25, for the whorled and monopodial examples, respectively. We directly present the two results for the autoencoded, noisy meshes. The reason for this is that the model naturally performed well on the base, synthetic meshes, a principle of machine learning as a whole. Despite using a held out randomization seed on the testing set---so as to avoid data snooping---the program still produced near-perfect results regardless. We believe it would be uninformative to simply conclude any results off of this simple result. As such, we encourage the reader to observe the results in Figures 24 and 25, where we display the autoencoded meshes. It is imperative to note that as there are holes and discontinuities in these inputs, it does not immediately hold that the model would perform well. Yet, despite these precautions, testing concludes that the model generalizes well outside of its learned space, evidence that it doesn't simply memorize models and that the autoregression applies well for plants. 

At this point, we have explored a narrative that encapsulates both a branch-only scheme as well as a full, plant mesh. For both, we find that autoregressive rigging automatically and intuitively creates accurate and robust plant skeletons, paving the way for future work. Thus, we dive deeper to uncover the technical intuitions behind our work, and conclude the experimental pipeline here.

\section{Discussion}
\fontsize{11}{13}\selectfont
The objective of this study was to evaluate whether an autoregressive rigging model originally developed for articulated objects could be adapted to reconstruct plant skeletal structures. Although UniRig was not designed with botanical data in mind, the experimental results demonstrate that it transfers remarkably well to plant meshes. 

With branch-only meshes, following fine-tuning on the synthetic L-system dataset, the model consistently produced skeletons that closely matched the underlying branching topology, achieving an estimated visual reconstruction accuracy of approximately 90\%. These findings suggest that the autoregressive formulation itself is not inherently limited to articulated objects, but instead generalizes to a broader class of hierarchical structures.

One explanation for this performance lies in the structural similarities between articulated skeletons and plants. While the biological functions of joints and branches differ substantially, both can be represented as rooted trees consisting of parent-child relationships. UniRig models skeleton generation as an autoregressive sequence prediction task, where each predicted joint is conditioned on the previously generated skeleton. Rather than attempting to infer the entire skeletal graph simultaneously, the model incrementally constructs the topology by exploiting contextual information accumulated during generation. This sequential prediction process aligns naturally with plant architecture, where branches recursively emerge from a single stem node to form increasingly complex hierarchical structures. Consequently, despite the absence of explicit botanical priors, the transformer is able to learn branching relationships directly from geometric patterns present in the training data.

Another contributing factor is the diversity of the procedurally generated dataset. Because the training data were generated using Lindenmayer systems, the model was exposed to a wide variety of underlying architectures while simultaneously receiving exact ground-truth skeletal annotations. Unlike manually annotated datasets, procedural generation guarantees perfect correspondence between mesh geometry and skeletal topology, eliminating annotation ambiguity during training. Furthermore, systematic variation in branching angle, branch length, branching order, and overall morphology encourages the model to learn general structural principles rather than memorizing individual plant shapes. This likely contributes to the model's ability to generalize across multiple archetypes despite their differing appearances.

The strong performance of UniRig can be attributed not only to its autoregressive
formulation but also to its skeletal \textit{representation}. Rather than directly regressing a skeletal graph, UniRig first converts the skeleton tree into a discrete token sequence via \emph{Skeleton Tree Tokenization} (STT). Bone tail coordinates, normalized to $[-1,1]^3$, are discretized into $D=256$ bins per axis via
$$M: x \mapsto \left\lfloor \frac{x+1}{2} \times D \right\rfloor \in \mathbb{Z}_D$$
so each joint $\mathcal{J}_i$ is represented by a discrete triple
$d_i = (dx_i, dy_i, dz_i)$, with average relative discretization error
$\mathcal{O}(1/D)$. A na\"ive depth-first traversal that concatenates
$(\text{parent coords}, \text{child coords})$ for every edge wastes tokens through
repeated parent coordinates and, empirically, induces repetitive degenerate sequences
at inference. UniRig instead performs a stack-based DFS in which a joint's coordinates
are emitted only once, with a dedicated \texttt{<branch\_token>} inserted precisely
when the traversal backtracks to a different parent than the previously emitted bone---i.e., exactly at branching points in the tree---and children at each joint are
visited in a canonical $(z,y,x)$ order, with $z$ being the vertical axis. Known skeleton templates (e.g.\ \texttt{<mixamo:body>}) are further collapsed into a single type token plus a flat coordinate run. This scheme reduces token count by roughly 30\%
relative to the na\"ive encoding on VRoid and Rig-XL while also improving joint
prediction accuracy at matched training budgets, and guarantees by construction
that any decoded sequence corresponds to a valid tree.

A GPT-style decoder-only transformer (OPT-125M) then predicts this sequence
autoregressively under a standard next-token-prediction objective,
$\mathcal{L}_{\text{NTP}} = -\sum_{t=1}^{T} \log P(s_t \mid s_1, \dots, s_{t-1}, \mathcal{F}_G)$,
where $\mathcal{F}_G$ is a geometric embedding produced by a randomly-initialized
3DShape2Vecset encoder over a 65{,}536-point surface sample with normals. $\mathcal{F}_G$
is \emph{prepended} to the token sequence rather than injected via a dedicated
cross-attention pathway, so mesh conditioning at every generation step arises from
the transformer's ordinary causal self-attention over this prefix.

This formulation is a particularly good structural fit for plant architecture, for
several concrete reasons. First, the token-efficiency gain from replacing per-edge parent repetition with sparse \texttt{<branch\_token>} insertion scales with how unbalanced the tree's branching factor is: a plant skeleton is dominated by long unbranching stem and petiole runs punctuated by comparatively rare forks. The 27--30\% token reduction reported for VRoid and Rig-XL---datasets of humanoid and quadruped rigs with moderate branching factor throughout the limbs---should be a conservative lower bound for plant meshes, where the ratio of chain tokens to branch tokens is considerably higher. Second, the $\mathcal{O}(1/D)$ discretization error is well matched to plant geometry. Branch diameter and curvature vary continuously and coarsely along a stem, without the sub-millimeter joint precision required at, for example, humanoid finger or facial rigs, so the fixed 256-bin resolution is unlikely to be a binding accuracy constraint for skeletal placement in the way it might be for articulated extremities. Third, the canonical $(z,y,x)$ child-sorting rule imposes a soft inductive bias toward emitting more vertically dominant children before laterals, which loosely mirrors apical dominance in real plant growth. Fourth, because $\mathcal{F}_G$ is prepended as a prefix rather than attended to locally, the decision to emit a \texttt{<branch\_token>} at any given step can in principle be informed by global mesh context (e.g., overall canopy silhouette or plant habit) rather than purely local junction geometry, which matters because whether a given point along a stem is a branch point is often only disambiguated by the plant's broader morphology rather than by the immediate mesh neighborhood.

This last point, however, is also where the fit breaks down empirically rather
than architecturally. Because \texttt{<branch\_token>} is comparatively rare in the
token distribution we found in earlier diagnostic work
that constrained top-$k$ sampling ($k=5$) at inference could exclude
\texttt{token\_id\_branch} from the candidate pool entirely, collapsing generation
into near-linear chains despite the tokenizer's capacity to represent branching
losslessly. The theoretical fit between STT and plant topology is therefore
necessary but not limiting. Moreover, the reported reconstruction accuracy is currently based on visual inspection rather than standardized quantitative metrics. While qualitative evaluation provides an intuitive assessment of skeletal plausibility, future work would benefit from objective measures such as graph edit distance, branch correspondence accuracy, and root-to-leaf path similarity. Such metrics would allow direct comparison with existing plant skeletonization algorithms and provide a more rigorous evaluation of reconstruction quality against existing schemes.

Another thing to note is that reconstruction errors occur primarily in densely branched regions or highly ambiguous branching (as with thicker branches), where nearby branch endpoints are occasionally merged or incorrectly connected. A possible explanation is that the tokenization or detokenization process introduces geometric ambiguity when multiple mesh endpoints occupy similar spatial locations. If confirmed, refining the skeletal representation or incorporating stronger topological constraints during decoding may improve reconstruction accuracy without substantial architectural changes.

\subsection{Proposition and Next Steps}
This project doesn't only have implications for static representations of plants. Rather, the most widespread application of rigging is for animation, or movement. For vegetation, this has directly ecological application, much of which has already been detailed above. However, only tracking the skeleton would be insufficient for such a task. As such, we need to find unique ways in which to apply skinning weights to plants, as they do not have a blanketing surface layer as with articulated characters. 

Thus, a question that we raised during the course of the project was: how do we segment the branches and the leaves? More specifically, UniRig already has internal functionality that carries unique labels for different parts of the body. For instance, the bones are tokenized differently than the skin, treated as a movement-based weight. More concretely, during the tokenization step, UniRig detects and assigns token labels for different types of appendages. For instance there's a token called \texttt{<mixamo>} for the object hull itself, as well as specifications for \texttt{<mixamo:body>} and \texttt{<mixamo:hand>}. Here, we note that one important observation would be the existence criteria for each of these tokens. We observed that having the \textit{hand} token implied that the \textit{body} subtoken existed, but having a \textit{body} token did not imply that the \textit{hand} subtoken existed. Thinking intuitively about plant structures, it is clear that for vegetation, we had an issue of segmenting not only the \textit{blade} but also the attaching point to the branch, or the \textit{petiole stem}. Furthermore, we argue from an ecological standpoint that the existence of the petiole implies the existence of the blade, but the existence of the blade does not imply the existence of the petiole. For instance, take a pine needle versus an oak leaf. We assert that the body of the leaf is the entirety of the needle, which attaches directly to the branch itself, albeit pine needles are often dense in nature. On the other hand, the oak leaf has a small, thin stem---which is the petiole---that serves as the connector between the rigid branch and the flimsy leaf. Previously, skinning weights could be applied directly and uniformly across most articulated characters. However, one can see that the complexity of the situation arises when we concern ourselves with plants instead. Had we had more time, we would implement such a tokenization scheme. 

Another limitation to our current work is the lack of variety in the lesser attachments to vegetation. For instance, one the most prominent parts of a plant is a \textit{flower}, and oftentimes what draws humans to certain types of plants in the first place. On the same note, fruit proposes a unique attachment challenge separate from leaves, due to their variable stem. Continuing off of the proposal above, future work could be aimed towards developing unique token schemes for different kinds of plant structures. That way, the program that we develop can truly be distributed globally and work well in any environment, not just with basic trees or bushes. There must be other ways to segment these minor appendages. For instance, consider a flower. Like a leaf, it has two parts, the nexus and the petals usually. As such, it is impossible to have a flower without a nexus, since the petals all have to connect something that stores pollen. Hence, we find that we could introduce \texttt{<flower>} tokens, which have \texttt{nexus} and \texttt{petal} subtokens. Similarly to above, \texttt{petal} implies \texttt{nexus}, but the reverse does not have to hold (for instance, with apetalous flowers). 

Moreover, another characterization that UniRig carries internally is the \texttt{<spring\_bone>} token, simulating the physics of a literal tension spring. We argue that this kind of stiffness metric is present in the petiole of a leaf, even for some branches as well. While most thick parts of a plant are not prone to movement under external force, such as a tree trunk, some thinner parts oscillate under weather conditions such as wind. Leaves sway, anchored by their petioles to the branch. From this, small shockwaves pervade throughout thinner branches, causing them to oscillate lightly as well. Hence, having some form of a stiffness metric would increase the efficacy of dynamic---or motion-based---inference for plant ecology by a considerable amount. The reason why this customization would be important is due to the huge differences in plant movement, not just down to scalar weights such as with skin. Skin behaves as smooth surface. Foliage has so many different surface components that differ between species, making the problem far more complicated. For instance, even the aforementioned, lacking pieces to our considerations (fruits, flowers, buds) all must be rigged in distinct ways from each other. Simply having a single weight for the surface would not suffice. As such, we propose such the stiffness encoding scheme, which would allow for a realistic approach in motion simulation, being able to customize movement of both the bones and the attached minor appendages.

\section{Conclusion}
\fontsize{11}{13}\selectfont
Our findings demonstrate that autoregressive rigging model such as UniRig, while limited in their zero-shot capacity to reconstruct plant skeletal structure, can be brought to strong plant skeletal reconstruction through targeted optimization. Initial diagnosis revealed clear failure modes: branching collapse under default sampling constraints, an encoder with poor sensitivity to structural variation in plant geometry, and instability introduced by naive inference-time interventions such as forced root-position injection. Addressing these issues iteratively, rather than through a single fine-tuning pass, proved essential. Successive rounds of fine-tuning allowed the model to progressively recover branching topology and adapt to the structural idiosyncrasies of plant meshes, including the zero-thickness, mesh-normal-dependent geometry of leaves that posed particular difficulty for the frozen 3DShape2Vecset encoder.

Critically, the resulting model did not merely memorize the synthetic archetypes used in training. It generalized across a diverse range of plant forms, varying in branching complexity, fork angle, and the presence of foliage. This suggests that the fine-tuning process induced a more structurally grounded representation of plant topology rather than a narrow overfit to our procedural dataset. Furthermore, the successful extension to leafy specimens is particularly notable, since leaf geometry stresses exactly the encoder weaknesses we identified early in this work, and its inclusion suggests that our fine-tuning approach meaningfully addressed those underlying representational limitations rather than merely patching symptoms at the sampling stage.

These results carry several implications for future work and widespread application. First, they suggest that autoregressive rigging architectures originally designed for rigid or humanoid mesh domains naturally parallel to biological, branching structures. The gap is substantially one of training data and encoder adaptation rather than architectural incompatibility. This opens the door to extending the approach to a broader taxonomy of plant species, including those with more irregular or non-self-similar growth patterns than the L-system-derived archetypes used here. Second, our encoder-level findings point toward a promising direction of jointly fine-tuning the mesh encoder alongside the autoregressive decoder, rather than treating it as frozen, which may further improve sensitivity to fine-grained structural variation. Finally, robust automated plant rigging has practical downstream value for procedural content generation, digital twins of real vegetation, and biomechanical simulation of plant growth and movement — domains where manual rigging is currently a significant bottleneck. We view this work as a powerful demonstration that autoregressive rigging can be extended productively into plant ecology, with substantial room remaining to explore architectural and data-driven refinements.

\vspace{0.5cm}

\small
\textbf{Conflict of Interest:} The authors declare no conflict of interest.

\textbf{Author Contributions:} Author 1 contributed to conceptualization, methodology development, testing, package development and deployment, and full writing/editing of this manuscript. Authors 2 and 3 contributed to fine-tuning the model and providing the technical environment for implementation, as well as development of the framework used. 

\textbf{Funding:} This research received no external funding.

\textbf{Ethical Statement:} This study follows ethical and academic integrity guidelines.

\balance
\fontsize{11}{13}\selectfont
\printbibliography

@article{ref1,
  author  = {Chaudhury, A. and Godin, C.},
  title   = {Skeletonization of Plant Point Cloud Data Using Stochastic Optimization            Framework},
  journal = {Frontiers in Plant Science},
  year    = {2020},
  volume  = {11},
  pages   = {773}
}

@article{ref2,
  author  = {Wu, S. and Wen, W. and Xiao, B. and Guo, X. and Du, J. and Wang, C. and            Wang, Y.},
  title   = {An Accurate Skeleton Extraction Approach From 3D Point Clouds of Maize             Plants},
  journal = {Frontiers in Plant Science},
  year    = {2019},
  volume  = {10},
  pages   = {248}
}

@article{ref3,
  author  = {Krisanski, S. and others},
  title   = {Smart-Tree: Neural Medial Axis Approximation of Point Clouds for 3D Tree           Skeletonization},
  journal = {arXiv preprint arXiv:2303.11560},
  year    = {2023}
}

@article{ref4,
  author  = {Zhang, J.-P. and others},
  title   = {One Model to Rig Them All: Diverse Skeleton Rigging with UniRig},
  journal = {ACM Transactions on Graphics},
  year    = {2025},
  note    = {arXiv:2504.12451}
}

@article{ref5,
  author  = {Hu, M. and others},
  title   = {Skin Tokens: A Learned Compact Representation for Unified Autoregressive Rigging},
  journal = {arXiv preprint arXiv:2602.04805},
  year    = {2026}
}

@book{ref6,
  author    = {Prusinkiewicz, P. and Lindenmayer, A.},
  title     = {The Algorithmic Beauty of Plants},
  publisher = {Springer-Verlag},
  year      = {1990}
}

@article{ref7,
  author  = {Liu, X. and Santo, H. and Toda, Y. and Okura, F.},
  title   = {PlantPose: Universal Plant Skeleton Estimation via Tree-constrained                Graph Generation},
  journal = {arXiv preprint arXiv:2605.17773},
  year    = {2026}
}

@article{ref8,
  author  = {Monteiro, J. and others},
  title   = {3D Functional-Structural Plant Modelling for Agricultural Digital Twins:           A Domain Analysis},
  journal = {Computers and Electronics in Agriculture},
  year    = {2021},
  note    = {ScienceDirect, S016816992100257X}
}

@article{ref9,
  author  = {Yang, Y. and Shinoda, R. and Santo, H. and Okura F.},
  title   = {GaussianPlant: Structure-aligned Gaussian Splatting for 3D Reconstruction of Plants},
  journal = {IEEE Transactions on Pattern Analysis and Machine Intelligence},
  year    = {2025},
  note    = {https://arxiv.org/abs/2512.14087}
}

\end{document}